\pdfoutput=1

\documentclass[11pt]{article}
\usepackage{authblk}
\usepackage{acl}

\usepackage{times}
\usepackage{latexsym}

\usepackage[T1]{fontenc}

\usepackage[utf8]{inputenc}

\usepackage{microtype}

\usepackage{graphicx}
\usepackage{mathtools}

\usepackage{subcaption}
\usepackage{multirow}
\usepackage{array}
\usepackage{booktabs} 
\usepackage{amsmath}
\usepackage{amssymb}
\usepackage{bigdelim}

\title{Is the Elephant Flying? \\ Resolving Ambiguities in Text-to-Image Generative Models}

\author{Ninareh Mehrabi\textsuperscript{\rm *1}, Palash Goyal\textsuperscript{\rm 2}, Apurv Verma\textsuperscript{\rm 2}, Jwala Dhamala\textsuperscript{\rm 2},\\Varun Kumar\textsuperscript{\rm 2}, Qian Hu\textsuperscript{\rm 2}, Kai-Wei Chang\textsuperscript{\rm 2}, Richard Zemel\textsuperscript{\rm 2},\\Aram Galstyan\textsuperscript{\rm 2}, Rahul Gupta\textsuperscript{\rm 2}}

\affil{\textsuperscript{\rm 1}Information Sciences Institute, University of Southern California \\ \textsuperscript{\rm 2}Amazon Alexa AI-NU}

\begin{document}
\maketitle
\begingroup\def\thefootnote{*}\footnotetext{This work was done as an intern at Amazon Alexa.}\endgroup

\begin{abstract}
Natural language often contains ambiguities that can lead to misinterpretation and miscommunication. While humans can handle ambiguities effectively by asking clarifying questions and/or relying on contextual cues and common-sense knowledge, resolving ambiguities can be notoriously  hard for machines. In this work, we study ambiguities that arise in text-to-image generative models. We curate a benchmark dataset covering different types of ambiguities that occur in these systems.\footnote{Data will be publicly released upon publication.} We then propose a framework to mitigate ambiguities in the prompts given to the systems by soliciting clarifications from the user. Through automatic and human evaluations, we show the effectiveness of our framework in generating more faithful images aligned with human intention in the presence of ambiguities.
\end{abstract}
\section{Introduction}
Natural conversations contain inherent ambiguities due to potentially multiple interpretations of the same utterance. 
Different types of ambiguities can be attributed to  \textit{syntax}  (e.g., ``an elephant and a bird flying” — is the elephant flying?), \textit{semantics} (e.g., ``a picture of cricket” — is cricket referring to an insect or a game?), and  \textit{underspecification} (e.g., ``doctor talking to a nurse” — is the doctor/nurse male or female?). Ambiguities pose an important challenge for many natural language understanding tasks and have been studied extensively in the context of machine translation~\cite{stahlberg-kumar-2022-jam}, conversational question answering~\cite{guo2021abgcoqa}, and task-oriented dialogue systems~\cite{qian-etal-2022-database}, among others.

\begin{figure}[t]
    \centering
    \includegraphics[width=0.97\linewidth,trim=2.5cm 0.3cm 16cm 0cm,clip=true]{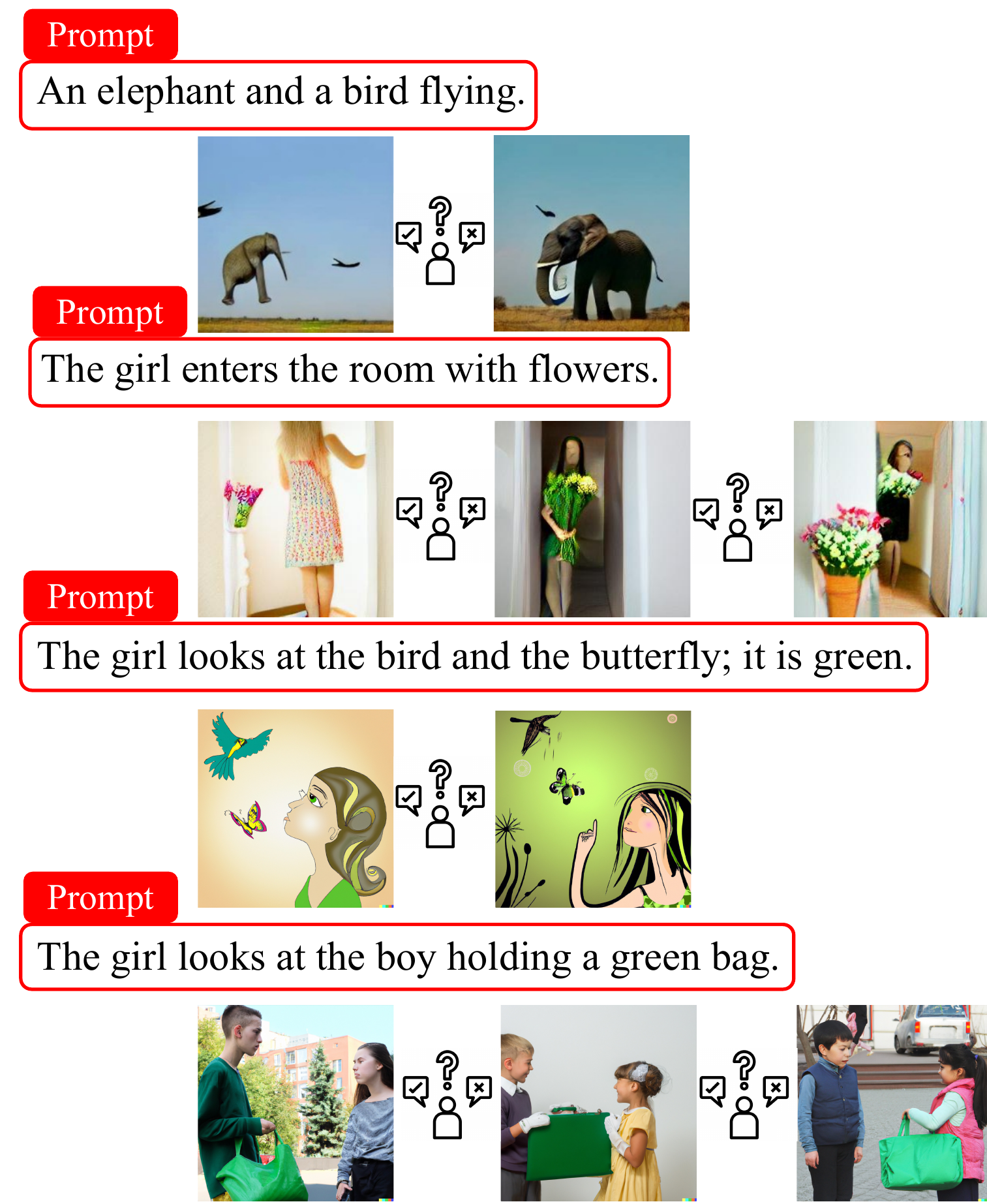}
    \vspace{-.07in}
    \caption{Examples of ambiguous prompts and corresponding generated images. The icon between the images depicts alternative interpretations. The images corresponding to first two prompts are generated by DALL-E Mega~\cite{Dayma_DALL}, and images corresponding to last two prompts are generated by OpenAI's DALL-E~\cite{ramesh2022hierarchical} models.}
    \label{fig:motivation}
    \vspace{-0.12in}
\end{figure}

\begin{figure*}[t]
    \centering
    \includegraphics[width=0.9\linewidth,trim=1cm 8cm 1.5cm 0.2cm,clip=true]{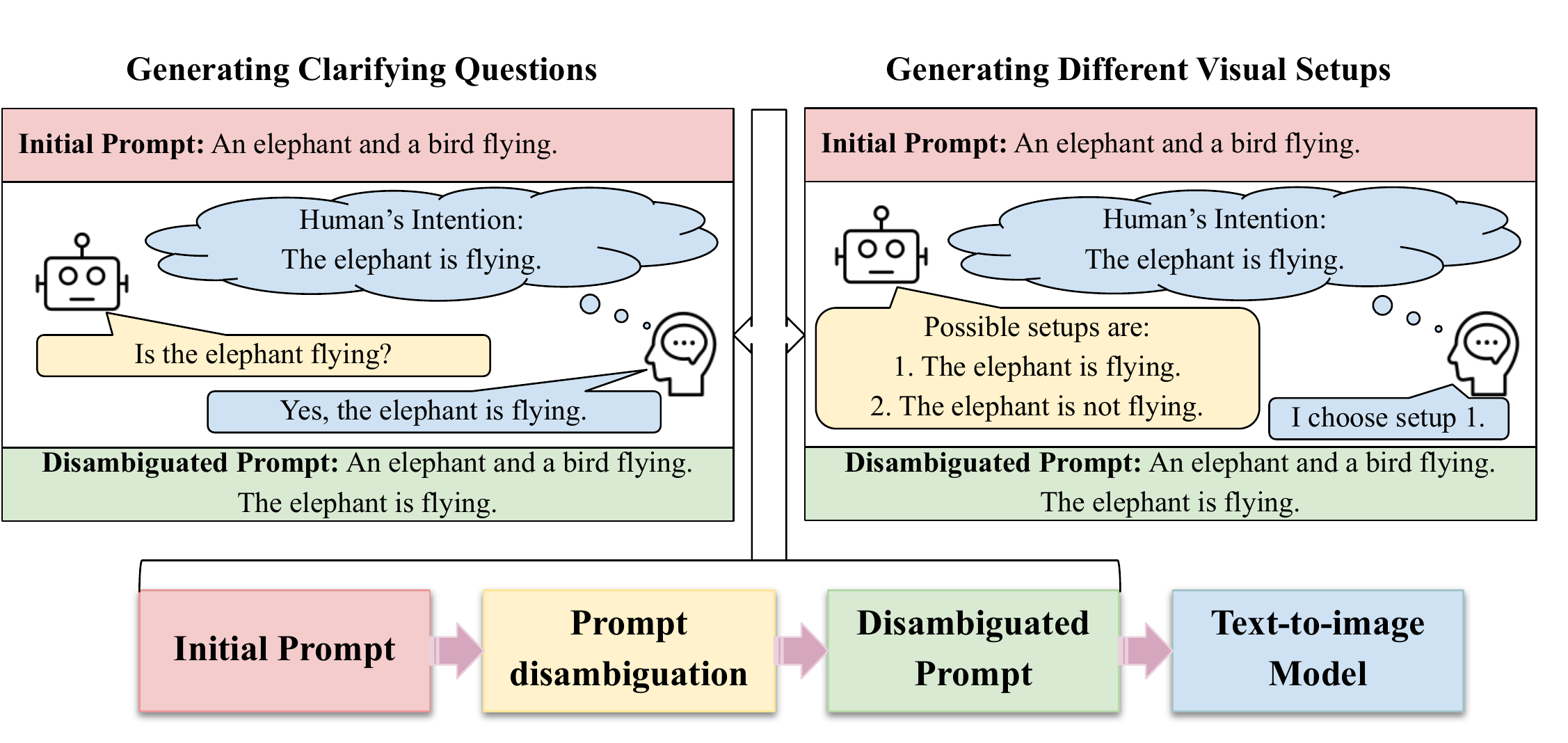}
    \vspace{-.07in}
    \caption{Our proposed disambiguation framework. The initial ambiguous prompt is disambiguated by either (1) the language model generating clarifying question which will be resolved through human provided answers, or (2) the language model generating different possible visual setups and human-agent choosing the desired setup. We define visual setups/interpretations as textual descriptions of different possible visual scenarios. The final disambiguated prompt will later be provided to the downstream text-to-image generative model.}
    \label{fig:framework}
\end{figure*}

In this paper, we study the effect of ambiguity in text-to-image generative models ~\cite{ramesh2021zero,ramesh2022hierarchical,saharia2022photorealistic,yu2022scaling} and demonstrate that ambiguous prompts provided to such models might result in undesired outcomes and poor user experience. In particular, ambiguities due to underspecification can lead to biased outcomes with possible implications on fairness of the underlying models (e.g., when prompted with ``doctor talking to a nurse”, the model might generate images with disproportionate number of male doctors and female nurses). We also propose a framework for mitigating ambiguities existing in prompts. We choose this setting to study ambiguity as visual scenes provide readily human-interpretable alternative understandings of text, thus helping in evaluating ambiguities as well as  mitigation strategies. Figure~\ref{fig:motivation} illustrates some sample prompts and corresponding outputs from the state-of-the-art text-to-image model, DALL-E. We observe that ambiguity in prompts confuses the model resulting in a diverse set of generated images. 

Humans tend to resolve ambiguities by asking clarifying questions, relying on other forms of modalities (such as vision), using contextual signals, and leveraging commonsense and/or an external source of knowledge~\cite{ACHIMOVA2022104862}. Inspired by  this observation, we propose a new framework (Figure~\ref{fig:framework}) in which we incorporate a language model-based {\em prompt disambiguation filter} on top of the text-to-image generative models. This filter is capable of either asking clarifying questions or generating different possible setups which would later be resolved through human interactions. Ultimately, the disambiguation filter helps the text-to-image model to identify a single visual setup for image generation.

To better understand the weaknesses of current text-to-image generative models, and to evaluate the effectiveness of our proposed mitigation framework, we  curate a benchmark dataset consisting of ambiguous prompts covering different types of ambiguities that are especially relevant to text-to-image models. We also propose new automatic evaluation procedures to evaluate faithfulness of generations in text-to-image generative models and compare it with human evaluations. We perform various automatic as well as human evaluation experiments to further validate our claims. 

Overall, we make the following contributions: \\
1. We introduce Text-to-image Ambiguity Benchmark (TAB) dataset containing different types of ambiguous prompts along with different visual setups (Section~\ref{sec:bench}). \\
2. We propose new automatic evaluation procedures to evaluate ambiguity resolution in text-to-image models (Section~\ref{sec:eval_method}). \\
3. We propose a disambiguation framework which can be applied on any text-to-image model. We use our benchmark dataset and the metrics to evaluate multiple variations of DALL-E as well as our disambiguation framework (Section~\ref{sec:framework}).

\section{Overview of the Framework}
Given an ambiguous prompt from our curated benchmark dataset, our proposed framework allows to disambiguate the initial  prompt through clarifying signals obtained via human-AI interaction. These signals are obtained by utilizing a language model that engages with the human-agent. Specifically, the framework either (1) generates clarifying questions for a human-agent to provide clarifying answers; or  (2) generates different possible setups that can disambiguate the prompt allowing the human-agent to pick the appropriate setup that matches the human intention. The framework is illustrated in Figure~\ref{fig:framework}.

Notice that we could also incorporate commonsense reasoning methods to resolve ambiguities in our framework; however, this approach would introduce some issues: (1) There might be cases in which all the interpretations might be commonsensical, and in this case the only reasonable way to get the appropriate signal would be through interactions with human-agent. (2) Even if not all the cases are commonsensical, the user still may want the un-commonsensical interpretation in which case our framework would be more suitable.

\section{Benchmark Dataset}
\label{sec:bench}
\begin{table}
 \scalebox{0.83}{
\begin{tabular}{ p{0.2cm} p{5cm} p{0.5cm} p{0.01cm}}
 \toprule
& \textbf{Ambiguity Type} & \textbf{Count}\\
 \midrule
 \parbox[t]{2mm}{\multirow{6}{*}{\rotatebox[origin=c]{90}{Main Types}}}&
 Syntax Prepositional Phrase (PP)&74& \rdelim\}{5}{3mm}[Linguistic]\\[0.5pt] 
 &Syntax Verb Phrase (VP)&243\\[0.5pt]
 &Syntax Conjunction&127 \\[0.5pt]
 &Discourse Anaphora& 21\\[0.5pt]
 &Discourse Ellipsis& 45\\[0.5pt]
 &Fairness& 355& \rdelim\}{1}{3mm}[Fairness]\\[0.5pt]
 \midrule
 \parbox[t]{2mm}{\multirow{1}{*}{\rotatebox[origin=c]{90}{Add}}} &
  Complex+combination+misc&335\\[4.0pt]
 \bottomrule
\end{tabular}}
\caption{Breakdown of our benchmark dataset TAB by ambiguity types. TAB consists of six types of ambiguities, including linguistic and fairness. We cover syntactic as well as discourse type ambiguities for linguistic type ambiguities. TAB also contains complex version for subset of the samples from the main type ambiguities with structurally more complex sentences, combination cases that combine fairness and linguistic type ambiguities, and some miscellaneous cases.}
\label{table:benchmark}
\end{table}
To develop Text-to-image Ambiguity Benchmark (TAB) dataset, we extend and modify the LAVA corpus~\cite{berzak-etal-2015-see}. The original LAVA corpus contains various types of ambiguous sentences that can be visually inspected/detected, along with their corresponding images/videos. We use the ambiguous prompts (templates) from LAVA and not the images-- as images in our case would be generated automatically by text-to-image generative models. The original LAVA corpus covers 237 ambiguous sentences (prompts) and 498 visual setups (possible interpretations for each ambiguous sentence). We expanded LAVA and curated TAB which covers 1200 ambiguous sentences (prompts) and 4690 visual setups. 

In addition to expanding LAVA, TAB also introduces a number of modifications, which include: (i) diversifying TAB to cover different objects, scenes, and scenarios, compared to LAVA, (ii) removing examples relevant to video domain from LAVA and only keeping examples relevant to static images in TAB, (iii) adding fairness prompts to TAB that cover different activities~\cite{zhao-etal-2017-men} and occupations~\cite{nadeem-etal-2021-stereoset} in which the identities of the individuals are ambiguous, (iv) adding more structurally complex sentences, and (v) additional labels for TAB (e.g., whether the visual setup or interpretation of an ambiguous sentence is commonsensical or not). More details of our modifications can be found in Appendix \ref{appendix:benchmakr-modifs}.

On a high level, TAB covers six main types of prompt ambiguities, including one fairness-related and five linguistic type ambiguities. In addition, we added some complex cases where we  took a sample of prompts from TAB and manually made structurally more complex version of each sentence. This modification is done in such a way that the ambiguity types as well as the meaning of a sentence are kept intact, while the structure of a sentence is made more complex through addition of more information, extra words, adverbs, and adjectives. Finally, we added some additional miscellaneous cases, which were not covered by six main types of ambiguities, as well as combination cases where we combined fairness and linguistic type ambiguities and made new variations from our existing prompts. Additional details can be found in Table~\ref{table:benchmark}. Appendix \ref{appendix:benchmakr-defs} also covers detailed definitions and examples for each of the ambiguities covered in TAB along with the dataset schema. Each of the elements in TAB including the labels and interpretations were cross-checked by two expert annotators.

\section{Disambiguation via Language Models}
\label{sec:framework}
After obtaining the initial ambiguous prompts from TAB, we utilize the capabilities of different language models to resolve existing ambiguities and obtain external disambiguation signals through human interaction using few-shot learning techniques, as discussed below.

\subsection{Method}
Our goal is to use few-shot learning utilizing a language model to disambiguate ambiguous prompts using human feedback. Toward this goal, we test two approaches. In the first approach, we give the language model examples of ambiguous prompts as model input and corresponding clarifying question as model output. We call this the ``clarifying question'' generation approach. In the second approach, we give the language model examples of ambiguous prompts as model input and descriptions of possible visual scenes as output. We call this the ``visual setup'' generation approach. The intuition is that the model will learn from these examples and generalize the ability to disambiguate other prompts during the inference time, as shown in Figure~\ref{fig:framework}. In our experiments, we provided each of the three language models that we tested (GPT-2~\cite{radford2019language}, GPT-neo~\cite{gpt-neo}, OPT~\cite{zhang2022opt}) one example from each of the main six types of ambiguities existing in TAB. Although these models may be small, it has been shown that small language models are also effective few-shot learners~\cite{schick2020s}. We then reported the ability of the language model in either generating one clarifying question per ambiguous prompt, multiple clarifying questions per prompt, or multiple possible visual interpretations per prompt. We consider each of these abilities as separate setups under which a language model will try to obtain external signals through human interaction.  The details of the few-shot examples given to the models as well as the details about each language model used can be found in Appendix~\ref{appendix:LM-experiments}.

After obtaining the results from the language model, we compare the generations to the ground truths provided in TAB. TAB is curated such that for each ambiguous prompt (e.g., ``\emph{An elephant and a bird flying}'') different possible visual interpretations that can be associated to a prompt are present (e.g., (1) ``\emph{The elephant is flying}'', (2) ``\emph{The elephant is not flying}'',). The existing visual interpretations in TAB serve as our ground truth for the ``visual setup'' generation approach. In addition to visual interpretations, TAB contains the question format of each of those interpretations (e.g., (1) ``\emph{Is the elephant flying?}'', (2) ``\emph{Is the elephant not flying}'') that serve as our ground truth for the ``clarifying question'' generation approach. Each ground truth was cross checked by two expert annotators to ensure the quality of the existing ground truth setups and questions in TAB. We acknowledge that for some cases, such as fairness ambiguities, it is impossible to cover all the ground truth interpretations and consider this as a limitation of our work. Please refer to the Appendix Section~\ref{sec:bench_appendix} where we demonstrate TAB dataset's schema for detailed explanation. For automatic evaluations, we report the BLEU and ROUGE scores by comparing the generations to the ground truth visual setups/clarifying questions we have provided in TAB.

To evaluate validity of the automatic metrics and their alignment with human-generated results, we perform the human interaction experiments with the language model, in which the human-agent provides disambiguation signals to the language model where appropriate. In this case, the response (disambiguation signal) is provided to the language model if the generated clarifying question is helpful in disambiguating the prompt according to the provided ground truth interpretation of the prompt from TAB and left unanswered otherwise. Similarly, for the setup where the language model generates multiple visual setups, the human-agent chooses the setup if it is appropriate given the ground truth interpretation and left unanswered otherwise. In this work, we refer to ground truth interpretation as human intention. Ultimately, the human-agent sees each ground truth interpretation from TAB and interacts with the system according to the given interpretation to satisfy their intention. Notice that TAB covers all the possible interpretations given an ambiguous prompt; thus, all the possibilities are considered by the human-agent interacting with the system one at a time which will be tracked and matched to its corresponding generated image later for the evaluation purposes.

These human interaction experiments will serve three purposes: 1. We can measure how aligned human evaluation is to automatic metrics to evaluate quality of generations by the utilized language models. 2. We will obtain the disambiguation signals that are later needed for comparing images generated by the text-to-image generative models using original ambiguous prompts vs images generated from prompts that are disambiguated given external signals. 3. Since the automatic metrics might not capture different variations of the ground truth labels, we perform human evaluations to ensure that the generations are still relevant.

\subsection{Results}
Here, we report the results for the case when the language models generate one clarifying question per given prompt. Additional results for the other experimental setups (generating different visual setups directly and multiple clarifying questions per prompt) can be found in Appendix~\ref{appendix:LM-results}. 
\begin{table*}
    \centering
    \scalebox{0.8}{
    \begin{tabular}{c | c c | c c |cc}
        \toprule
        & \multicolumn{2}{c|}{GPT-2} & \multicolumn{2}{c|}{GPT-neo} & \multicolumn{2}{c}{OPT}\\
        Ambiguity Type & BLEU $\uparrow$& ROUGE $\uparrow$& BLEU $\uparrow$& ROUGE $\uparrow$& BLEU $\uparrow$& ROUGE $\uparrow$\\
        \midrule
        Total Benchmark & 0.39&0.58&0.46&0.60&0.42&0.59\\
        Syntax Prepositional Phrase (PP) & 0.21&0.64& 0.06&0.63&0.22&0.65\\
        Syntax Verb Phrase (VP) & 0.60&0.81& 0.75&0.84&0.67&0.83\\
        Syntax Conjunction & 0.17&0.63&0.23&0.65&0.06&0.56\\
        Discourse Anaphora & 0.30&0.69&0.19&0.60&0.74&0.83\\
        Discourse Ellipsis & 0.48&0.69&0.22&0.47&0.55&0.75\\
        Fairness & 0.36&0.55&0.60&0.59&0.50&0.58\\
        \bottomrule
    \end{tabular}}
    \caption{BLEU and ROUGE scores obtained by different LMs on generating a clarifying question in 6-shot setup given an ambiguous prompt. $\uparrow$ indicates that higher values are desired. Scores are reported on a 0-1 scale.}
    \label{results:LM}
\end{table*}

From results in Table~\ref{results:LM}, we observe that, when provided with an ambiguous prompt, the language models have reasonable ability to generalize and generate good quality clarifying questions compared to our ground truth  according to BLEU ($\sim 0.40$) and ROUGE ($\sim 0.60$) metrics. Similar to the clarifying question generation results, the models were able to generate good quality results with reasonable BLEU and ROUGE scores for the visual setups generations. However, we note that better scores were obtained when language models were generating clarifying questions than when directly generating different possible visual setups. This suggests that the task of directly generating multiple setups given a few examples might be slightly harder for these models than generating clarifying questions given an ambiguous prompt.

In addition to reporting overall results on our benchmark dataset, fine-grained results for the six different ambiguity types, as reported in Table~\ref{results:LM}, suggest that there exists disparity in how different ambiguities are handled by each of these language models. For instance, language models obtain higher BLEU and ROUGE scores on generating clarifying question for ambiguity type Syntax Verb Phrase (VP) than ambiguity type Syntax Propositional Phrase (PP). This suggests that some types of ambiguities are easier for the language models to resolve compared to the others. 

\begin{figure}[t]
    \centering
    \includegraphics[width=0.85\linewidth,trim=0cm 0.4cm 0cm 0cm,clip=true]{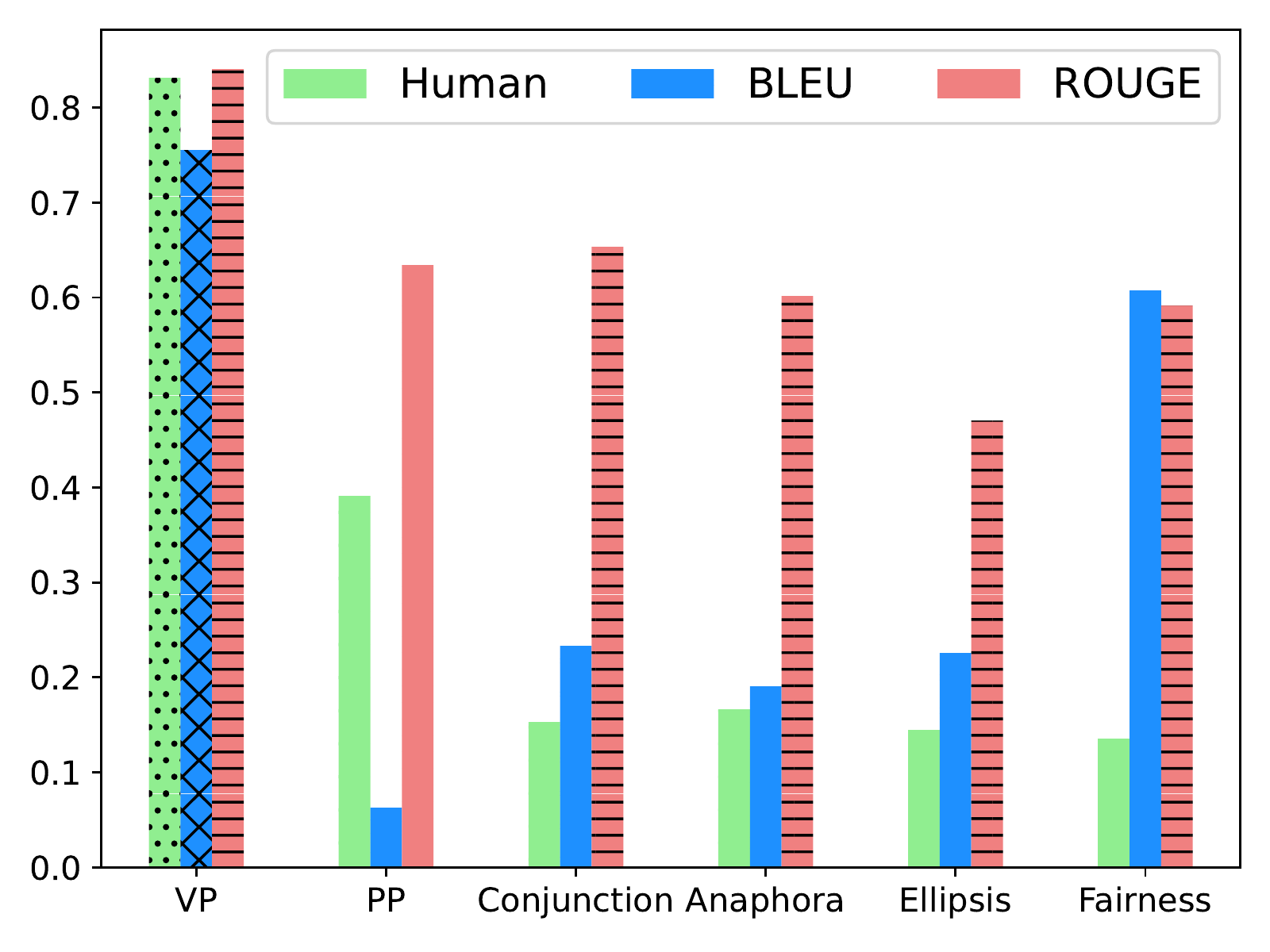}
    \vspace{-.07in}
    \caption{Percentage of generations by GPT-neo that were successful according to human-agent and its comparison to BLEU and ROUGE automatic metrics.}
    \label{fig:humanevals}
\end{figure}

Our results from our human interaction experiments are shown in Figure~\ref{fig:humanevals}. In these results, we report the percentage of generations by the GPT-neo language model that deemed successful according to the human-agent interacting with this model. In other words, we report the percentage of generations that an answer was provided by the human-agent and was not left unanswered. An unanswered generation would mean that the generation was not helpful for disambiguating the prompt and thus left unanswered by the human-agent. We acknowledge that this would subjectively depend on how well the human agent understood the disambiguation question generated by the model. However, based on manual inspections, the generated questions are simple and do not contain risks of misunderstandings. We perform this experiment on GPT-neo model as we obtained best automatic results for this model. From our human evaluation results, we report the agreement between automatic metrics and human evaluation results with Pearson correlation across ambiguity types between human and ROUGE score of 0.863 and between human and BLEU score of 0.546. We use Pearson correlation as it is widely used to report agreement between human and automated evaluations. This agreement can be observed from relative rankings amongst different ambiguity types in Figure~\ref{fig:humanevals}. The interpretation of these results is that if, according to ROUGE score, the model generates better results for one set of ambiguity compared to another (e.g., PP vs Anaphora), the same holds for human evaluation results. Similar results for generating visual setups are reported in Appendix ~\ref{appendix:LM-results}.

\paragraph{Ablation Studies} We perform several ablation studies to demonstrate the effect of the number of few-shot examples provided to the model on its performance as well as model's ability to resolve ambiguities for complex vs simple sentence structures. We will summarize our main findings here in the main text and refer the reader to Appendix~\ref{appendix:LM-results} for detailed results. We perform experiments in which, for a given ambiguity type we vary the amount of few-shot examples provided to the model and report model's performance on resolving the specific type of ambiguity,  as well as its generalization ability in resolving other ambiguity types. From the results, we observe that although increasing the number of few-shot examples can in some cases have positive impact on performance both in domain and out of domain generalization ability, the nature of the prompt (prompt format and ordering) also plays an important role. Our results also match the previous findings in~\cite{zhao2021calibrate} in which authors studied the effect of few-shot examples provided to language models to perform various tasks and report findings similar to ours. For a detailed discussion on this matter, we refer the reader to Appendix~\ref{appendix:LM-results}.

In a different experimental setting, we also compared the performance disparities between language models' ability in resolving existing ambiguities in simple sentences vs similar sentences with more complex sentence structures that we curated in TAB. As mentioned before, for a sample of existing sentences in TAB, we manually created more complex version of those sentences such that their meaning and ambiguity are kept intact,  while their structure is made more complex via injection of additional information, words, adjectives, and adverbs. Our results indicate that language models have lower performance for complex sentence structures compared to their simple sentence counterparts, which is expected. Those results and detailed discussion can be found in Appendix ~\ref{appendix:LM-results}. Qualitative examples as well as few-shot prompts provided to our models can be found in Appendix~\ref{appendix:LM-experiments}.

\section{Text to Image Disambiguation}
\begin{figure}[t]
    \centering
    \includegraphics[width=\linewidth,trim=2.5cm 2.3cm 5cm 4.5cm,clip=true]{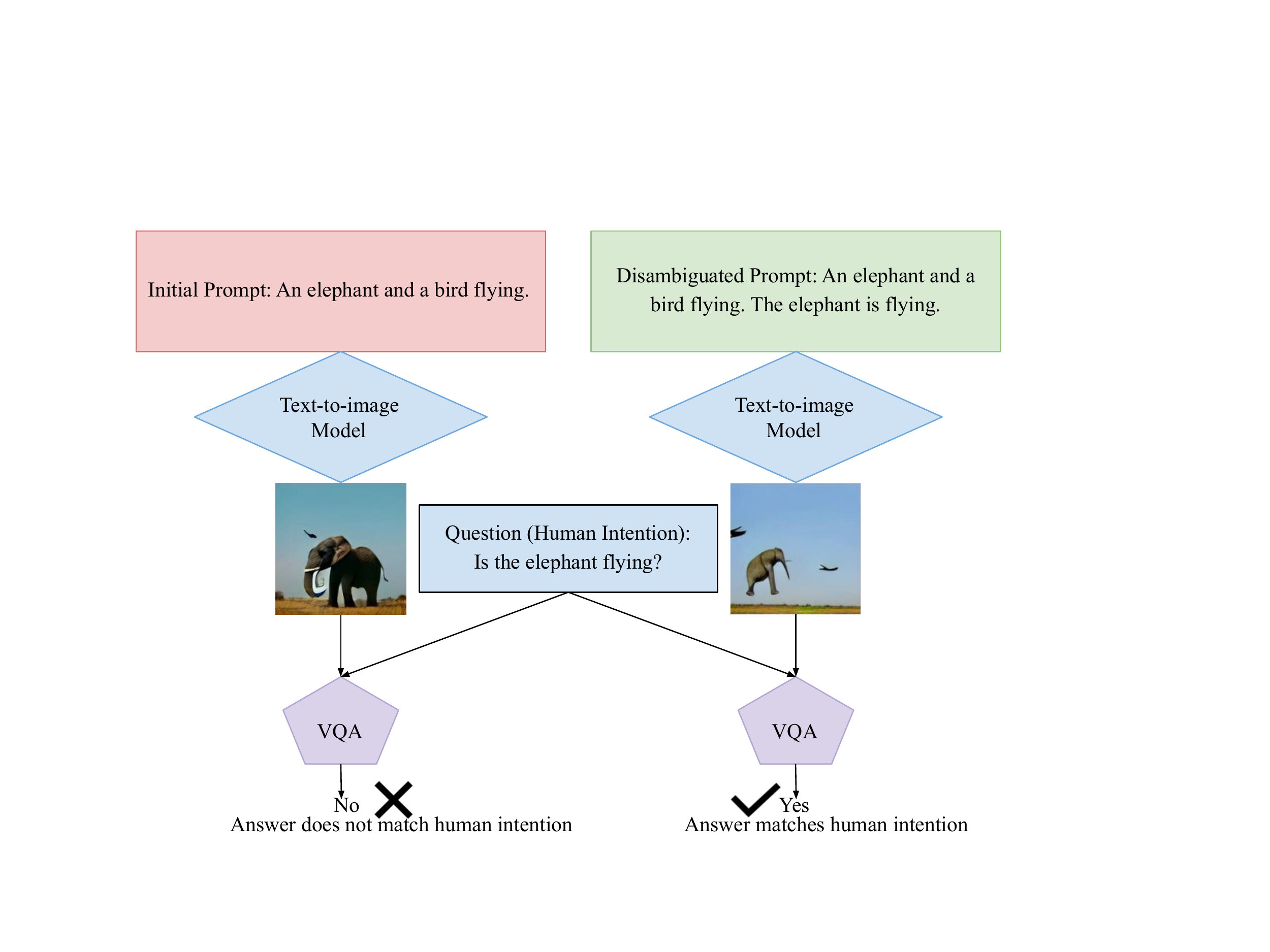}
    \caption{Our automatic evaluation framework using a VQA model. Given an ambiguous and a disambiguated prompt, we compare the generations made by the text-to-image model according to a VQA model. As inputs to the VQA model, we provide the human intention in the question format as well as the generated images from each prompt.}
    \label{fig:vqa}
\end{figure}
To evaluate the effectiveness of our framework and prompt disambiguation in allowing text-to-image generative models to generate more faithful images aligned with human intention, we compared generated images from OpenAI's DALL-E model~\cite{ramesh2022hierarchical} as well as DALL-E Mega~\cite{Dayma_DALL} using disambiguated prompts vs the original ambiguous prompts.

\subsection{Method}
\label{sec:eval_method}
\begin{figure*}
\centering
\begin{subfigure}[b]{0.37\textwidth}
\includegraphics[width=\textwidth,trim=0cm 0cm 0cm 0cm,clip=true]{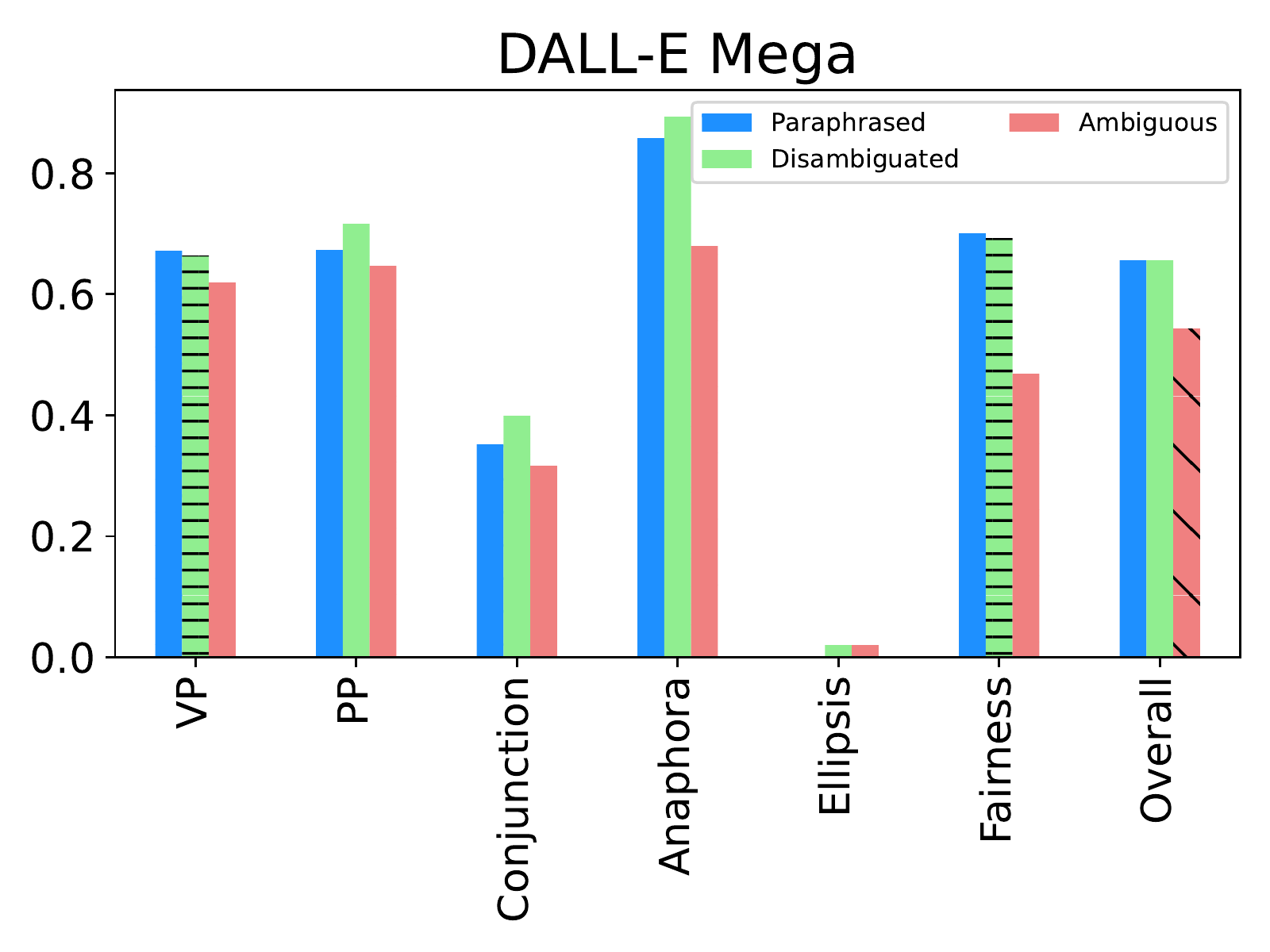}

\end{subfigure}
\begin{subfigure}[b]{0.37\textwidth}
\includegraphics[width=\textwidth,trim=0cm 0cm 0cm 0cm,clip=true]{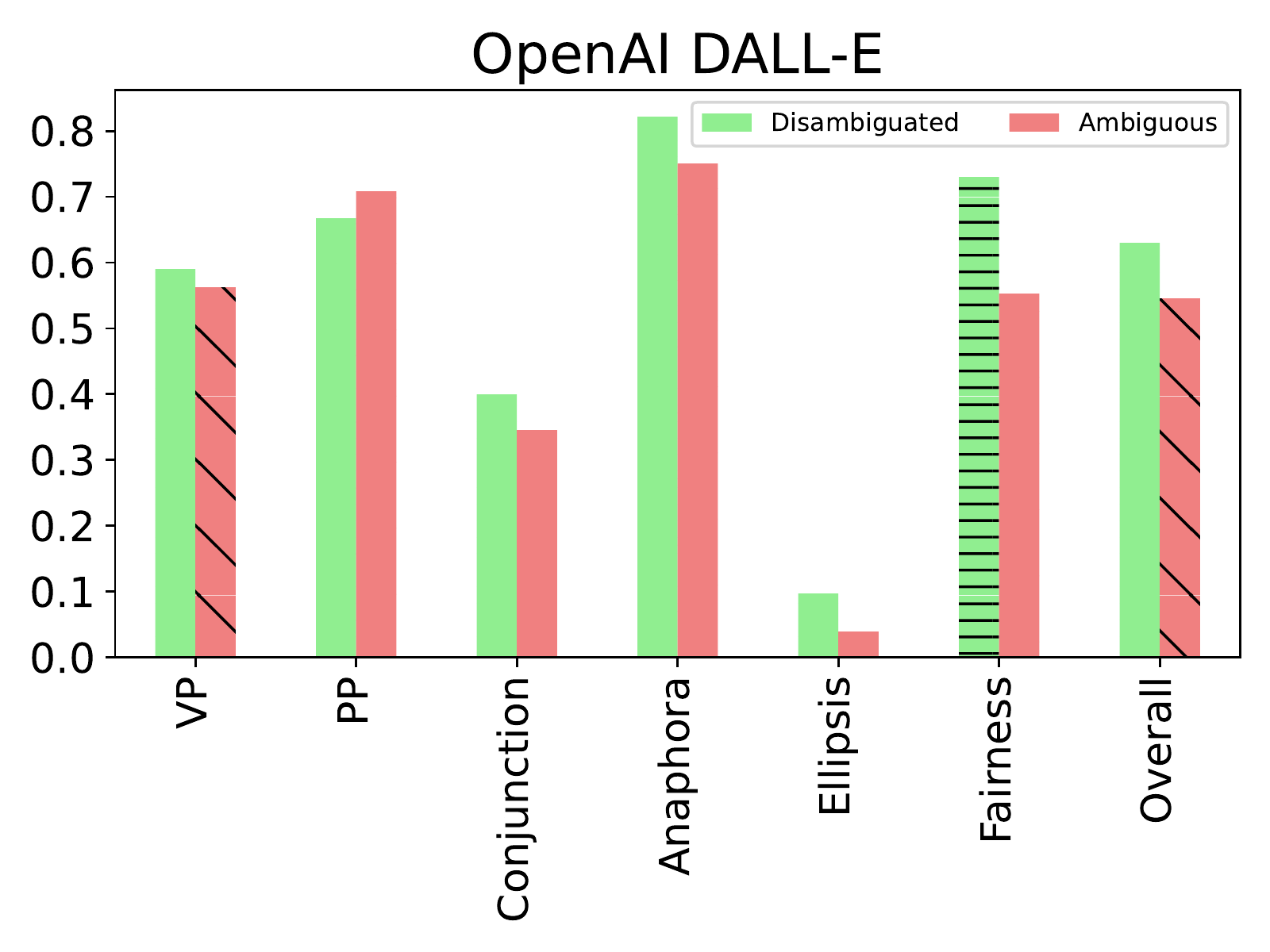}

\end{subfigure}
\caption{Percentages of faithful image generations by DALL-E Mega and OpenAI's DALL-E according to automatic evaluation using a VQA model.}
\label{fig:vqa-auto}
\end{figure*}

\begin{figure*}
\centering
\begin{subfigure}[b]{0.37\textwidth}
\includegraphics[width=\textwidth,trim=0cm 0cm 0cm 0cm,clip=true]{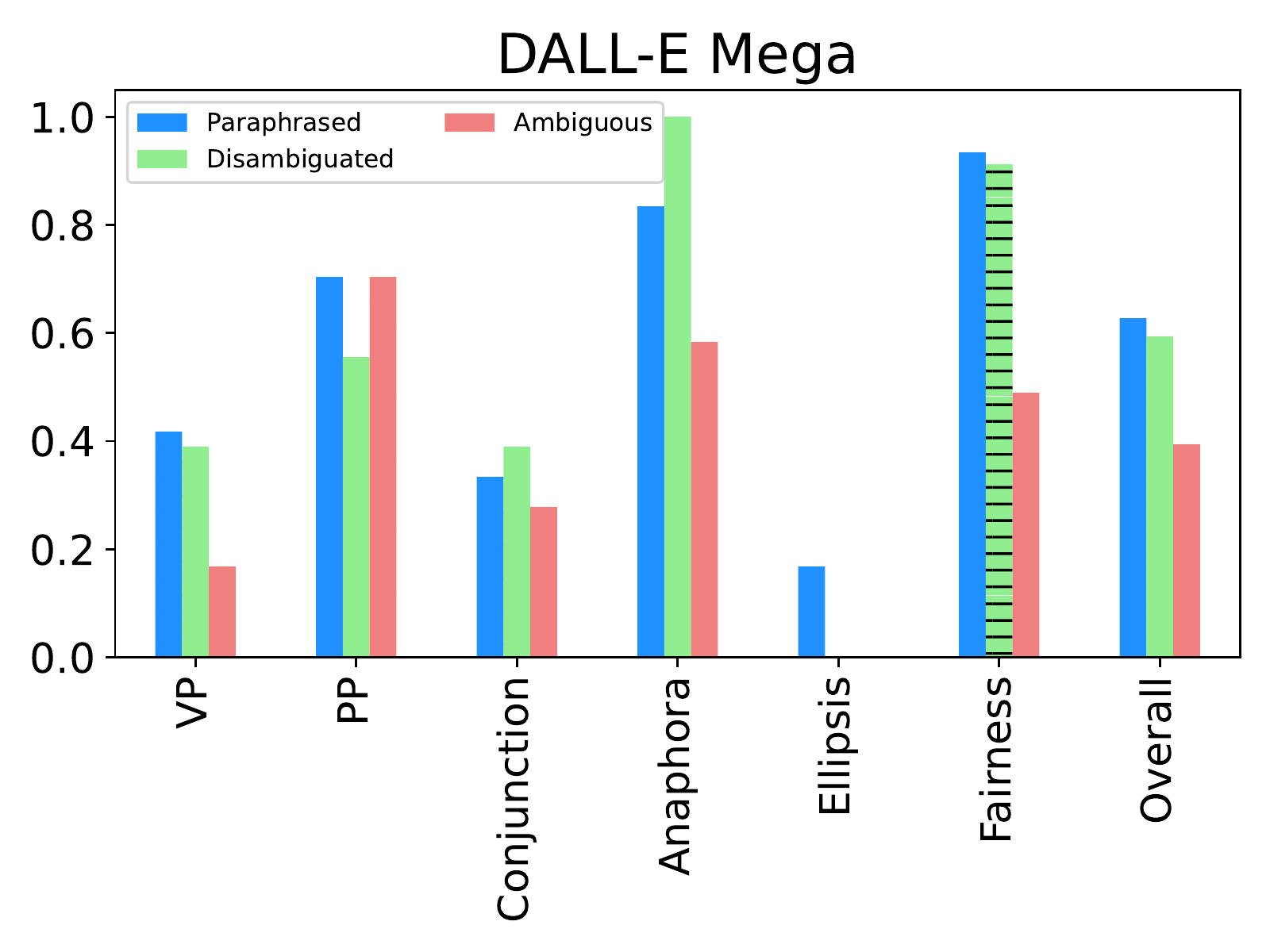}

\end{subfigure}
\begin{subfigure}[b]{0.37\textwidth}
\includegraphics[width=\textwidth,trim=0cm 0cm 0cm 0cm,clip=true]{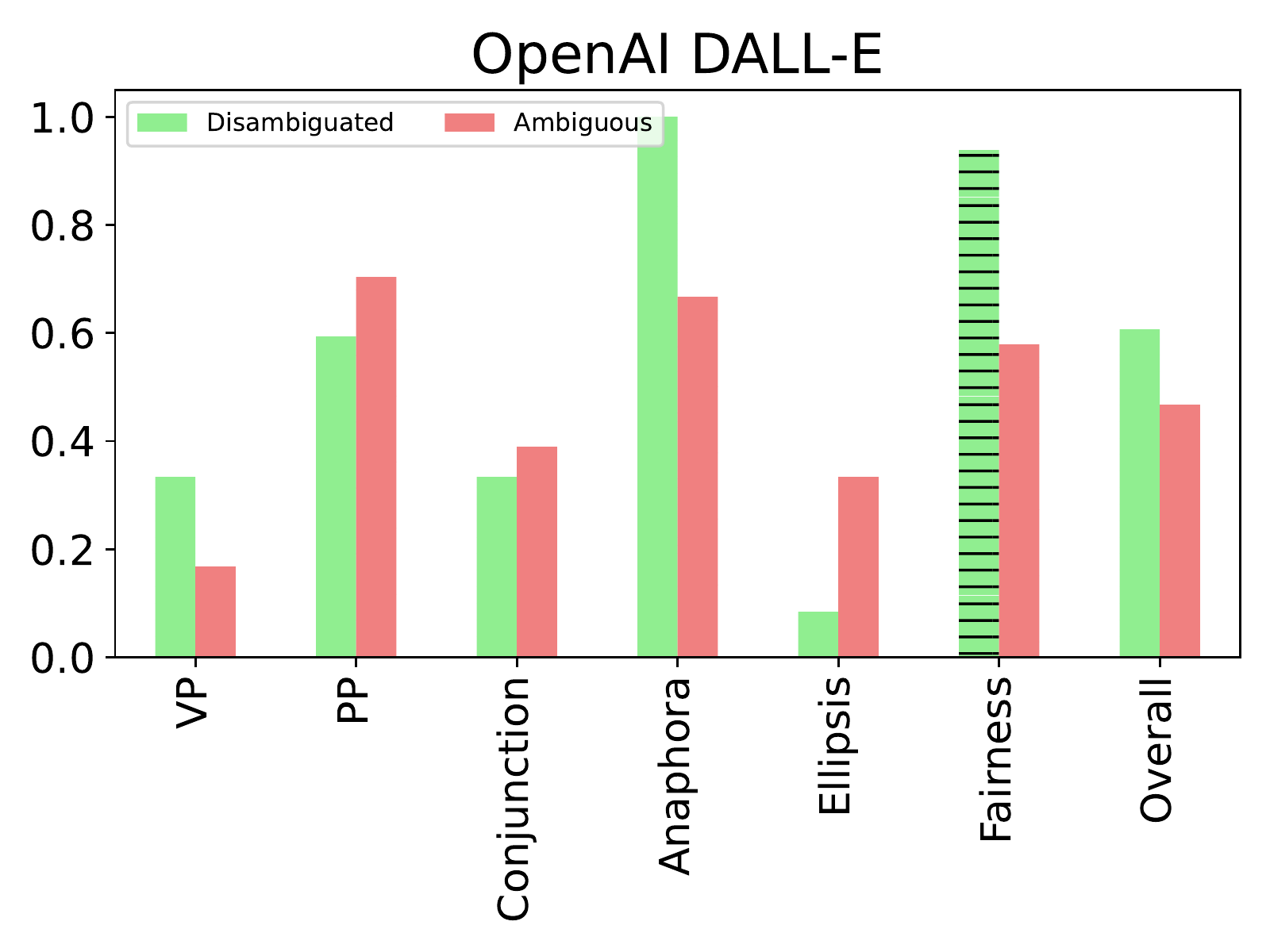}

\end{subfigure}
\caption{Percentage of faithful generations by DALL-E Mega and OpenAI's DALL-E from human evaluations. Note that for human evaluations, we use a subset of 400 generated images (more details in Appendix ~\ref{appendix:text-to-image}).}
\label{fig:mturk-eval}
\end{figure*}
After obtaining the disambiguation signals through language model vs human interactions from previous steps of our framework (either in the format of a human answering the clarifying questions raised by the language model or a human picking one of the possible visual setups generated by the model), we concatenated this external disambiguation signal to the original ambiguous prompt. To observe the effectiveness of this disambiguation framework in enabling text-to-image models to generate more faithful images, we compared the generations given the original ambiguous prompts to the ones that were disambiguated. For each prompt, four images were generated by OpenAI's DALL-E~\cite{ramesh2022hierarchical} and DALL-E Mega~\cite{Dayma_DALL} models. Overall, we generated and studied over 15k images for different setups, scenarios, and models. More details and statistics on prompts and images can be found in Appendix~\ref{appendix:text-to-image}.

After obtaining generated images per prompt from each of these models, we automatically evaluate faithfullness of generations  as well as effectiveness of our proposed framework by using the VILT Visual Question Answering (VQA) model~\cite{kim2021vilt}. TAB provides each image with human intention in the question format. We use both image and its corresponding question as inputs to the VQA model as shown in Figure~\ref{fig:vqa}. We then compare the results generated using the initial ambiguous prompts and the disambiguated prompts. Ideally, if the image aligns to human intention,  we would expect the VQA model to output a "Yes" as an answer to the question. Thus, we report the percentage of times the VQA model outputs a "Yes" as an answer as the percentage of faithful generations aligned with human intention amongst all the generations. 

Finally, to evaluate how reliable the proposed automatic evaluation method is in evaluating faithfulness of image generations, we perform human evaluations. Human evaluations serve two purposes. First, we want to evaluate whether our proposed automatic evaluation method using a VQA model is a reliable method to measure the faithfullness of image generation. Second, we want to evaluate the faithfulness of image generation after our proposed prompt-disambiguation using human judgement rather than automated metrics. The human evaluation experiments were performed on Amazon's mechanical turk platform. Overall, 400 images were annotated by mturk workers. Each image was annotated by three workers, so that we obtained in total 1200 annotations. More details about the mturk experiments along with our survey can be found in Appendix~\ref{appendix:text-to-image}.

Lastly, we explore the effect that paraphrasing of the disambiguated prompts can have on creating prompts more aligned with human intention, hence leading to more faithful image generation. For this purpose, we take all the disambiguated prompts obtained through our framework, which were concatenation of disambiguated signals provided by humans to the ambiguous prompts, and apply sentence paraphrasing model fine-tuned on BART over them~\cite{lewis-etal-2020-bart}. We then compare the results from providing the text-to-image model the ambiguous prompt vs the disambiguated prompt which was obtained from simple concatenation of human provided signal to the original prompt vs a paraphrased version of the disambiguated prompt from the previous step. We report the effect on whether this paraphrasing helps into the model generating more faithful images to human intention.
  
\subsection{Results}
\begin{figure*}[ht]
    \centering
    \includegraphics[width=\linewidth,trim=2cm 10.2cm 2cm 0cm,clip=true]{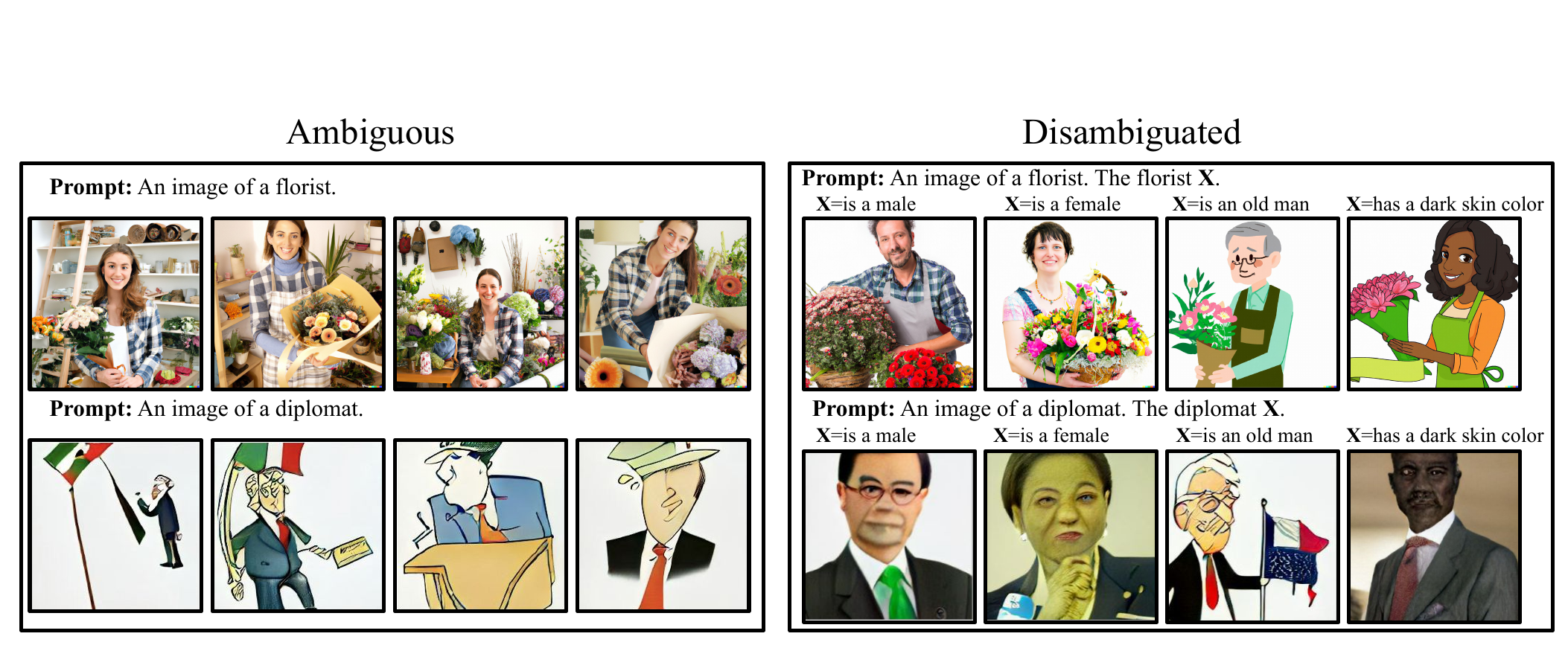}
    \vspace{-.07in}
    \caption{Fairness qualitative examples from OpenAI's DALL-E (top row) and DALL-E Mega (bottom row).}
    \label{fig:qualitative_fairness}
\end{figure*}

First, we demonstrate the effectiveness of proposed disambiguation framework into generating more faithful images that are better aligned with human intention according to human evaluations in OpenAI's DALL-E and DALL-E Mega models. Figure~\ref{fig:mturk-eval} shows the fraction of times the models generate faithful images. We observe that overall, disambiguation helps with faithful generations by improving the results from the baseline  that uses the original ambiguous prompts. Despite the overall positive impact of disambiguation, the fine-grained results in Figure~\ref{fig:mturk-eval} demonstrate that disambiguation might have adverse effect for some ambiguity types (e.g., PP type ambiguity). In addition, we observe that it is harder to generate faithful images that can even hit the random chance of generating the correct interpretations for some ambiguity types (e.g., Ellipsis) due to the complexity of the prompts in this ambiguity category for text-to-image generative models. Typically the Ellipsis category requires many entities being generated in the images in complex fashions that current text-to-image models are not capable of achieving it yet. In addition, we report the annotator agreement according to Fleiss Kappa~\cite{fleiss1971measuring} to be 0.86 which is a significant agreement. 

Second, we show similar results when using our proposed automatic evaluation method. We report Pearson correlation between human vs automatic evaluations to be 0.83 and 0.75 for DALL-E Mega and OpenAI's DALL-E respectively. Thus, the proposed automatic metric has reasonably high correlation with human annotators. For additional results on other setups as well as more evaluated image samples refer to Appendix~\ref{appendix:text-to-image-results}. The results are similar to findings reported in the main text.

Figure~\ref{fig:qualitative_fairness} demonstrates the effect that disambiguation has on generating more diverse images with implications on fairness. By specifying more specific identities associated to an individual and avoiding general ambiguous prompts, more diverse images can be generated. We believe that having a language model that is aware of these ambiguities and provides the user the opportunity to specify its intention more clearly can improve both user satisfaction as well as resulting these models to generate more diverse images with implications on fairness. More qualitative examples on linguistic cases can be found in Appendix~\ref{appendix:text-to-image-results}.

Finally, we report the impact that paraphrasing of the disambiguated prompts has on faithful image generation.  Figure~\ref{fig:vqa-auto} demonstrates that paraphrasing disambiguated prompts leads to slight and not significant improvement according to automatic VQA evaluation method using DALL-E Mega model and slightly more significant improvements according to human evaluations as shown in Figure~\ref{fig:mturk-eval}. We report these sets of results only on DALL-E Mega as most of the results follow similar patterns for both models (DALL-E Mega and OpenAI's DALL-E models).

\section{Related Work}
Resolving ambiguities in different NLP applications has been a prominent research direction due to its importance. For instance, word sense disambiguation is one of the areas in NLP that has gained significant attention and numerous works have been proposed in this regards~\cite{wang-wang-2021-word}. Resolving ambiguities in question answering~\cite{min-etal-2020-ambigqa}, conversational question answering~\cite{guo2021abgcoqa}, and task-oriented dialogue systems~\cite{qian-etal-2022-database} has also been previously studied.  Ambiguity resolution has also been studied in multi-modal applications, such as multi-modal machine translation~\cite{li2022visa} or matching images or videos to disambiguated interpretation of a sentence~\cite{berzak-etal-2015-see}. Despite those recent efforts, not much attention has been paid to ambiguities in text-to-image generative models. On the other hand, the growing popularity of those models, both in academic and non-academic circles, make it imperatives to better understand potential issues with those systems due to language ambiguity. In this paper we have identified and addressed some of those issues. We hope that our work will inspire future effort on this important problem.
\section{Conclusion}
In this work, we studied the role of prompt ambiguity in text-to-image generative models and proposed a disambiguation framework for helping the model to generate  more faithful images that  are better aligned with user intention.  
We first curated a benchmark dataset consisting of different types of ambiguities. We then measured the ability of various language models to produce disambigutating signals through human interaction, by either generating clarifying questions or directly generating multiple possible visual setups utilizing concepts from few-shot learning. After obtaining the signals via language model vs human interaction and performing different automatic as well as human evaluations, we measured the faithfulness of image generations by text-to-image generative models providing ambiguous, disambiguated, as well as paraphrased disambiguated prompts to these systems. We performed various automatic as well as human evaluations to validate the effectiveness of the proposed prompt disambiguation framework. In this work, most of our focus was on resolving ambiguities and not so much its detection which can be another interesting avenue for future research.

\section*{Ethical Considerations} 
In this work, we studied and proposed solutions to resolve existing ambiguities in prompts given to text-to-image generative models. In addition to resolving ambiguities in prompts, this work not only framed and analyzed fairness from a new and different prospective, but it also resulted in more faithful image generations aligned with human intention. These aspects can contribute to numerous positive impacts to the research community. Not only one can generate more diverse images through disambiguating fairness type ambiguities, but our framework can also improve user satisfaction by generating aligned images to human's intention despite existing ambiguities in the provided prompts. Resolving ambiguities can also avoid spread of misinformation and development of fallacies. 

Despite the aforementioned positive impacts, we also acknowledge the limitations associated with this work. We acknowledge that our benchmark dataset is just a very small sample of different types of ambiguous prompts that can be provided to a system. In addition, for the fairness type ambiguities, we only consider gender (male vs female), skin color (dark vs light), and age (young vs old). We acknowledge that these are only a limited number of characteristics that can represent identity of an individual and that we do not cover all the cases possible. We agree that we might not cover all the cases possible; however, our intent was to showcase a few examples through our benchmark dataset (TAB) and highlight the existing flaws associated with these systems encountering ambiguous prompts. 

In our experiments, we also utilized human annotators. We made sure to provide appropriate guidelines with a proper compensation to our workers (around 12\$ per hour). We also utilized master workers based in the United States with proper expertise (completion of more than 1000 HITs with an acceptance rate above 85\%). In addition, we provided the workers the opportunity to raise any concerns about our task. Based on the feedback, we believe that the task as well as the pay was satisfactory to the workers.

We hope that our study can provide valuable insights to the research community with the positive implications out-weighting its limitations. We will also open-source our benchmark dataset upon publication for the community to benefit from our work. As future work, researchers can investigate and propose better alternatives than our proposed framework for resolving ambiguities in text-to-image generative models along with extension of our work to semantic ambiguities in addition to the ones studied in this paper. Our benchmark dataset can also serve as a valuable resource for research in commonsense reasoning studies in text-to-image generative models which was less explored in our current work. We provided information in our benchmark dataset (whether an interpretation is commonsensical or not) which can be accessible to interested researchers in this area.
\bibliography{acl_latex}
\clearpage
\appendix
\section*{Appendix}
In this appendix, we will include details that were left out from the main text of the paper due to space limitations including experimental setup details as well as additional results and discussions. We ran all the experiments on an AWS p3.2xlarge EC2 instance. 

\section{Details About Benchmark Dataset}
\label{sec:bench_appendix}
Here, we will first define each of the different types of ambiguities existing in our benchmark dataset (TAB) with a corresponding example. We will then list the details of the modifications along with the extensions made to the original LAVA~\cite{berzak-etal-2015-see} corpus to make TAB.

\subsection{Definitions}
\begin{table*}[!t]
\centering
\scalebox{0.55}{
\begin{tabular}{ c |c| c| c}
 \toprule
\textbf{Example}&\textbf{Visual Setups}&\textbf{Commonsensical or Uncommonsensical}&\textbf{Question Format of Visual Setup}\\
 \midrule
An elephant and a bird flying
&[the elephant is flying, the elephant is not flying]&[UCS,CS]&[is the elephant flying?, is the elephant not flying?]\\[0.5pt]
 \bottomrule
\end{tabular}}
 \caption{Dataset schema of our benchmark (TAB) along with one provided example. The example contains the ambiguous prompt. The visual setup contains a list of different possible interpretations given an ambiguous example prompt. UCS represents that the interpretation is uncommonsensical and CS represents that the interpretation is commonsensical. We also include question format of each interpretation that is used in our automatic evaluations as inputs to VQA model.}
\label{dataset-schema}
\end{table*}
\label{appendix:benchmakr-defs}
\textbf{Syntax Prepositional Phrase (PP):} For this type of syntactic  ambiguity, we borrowed the following template \textit{\textbf{NNP V DT [JJ] NN\textsubscript{1} IN DT [JJ] NN\textsubscript{2}}} from the LAVA corpus~\cite{berzak-etal-2015-see} to construct most of the cases in TAB. An example for this type of ambiguity can be: \textit{The girl approaches the shelf with a green plate}. It is possible that 1. the green plate is with the girl or 2. is on the shelf.\\
\textbf{Syntax Verb Phrase (VP):} For this type of syntactic ambiguity, we borrowed the following template \textit{\textbf{NNP\textsubscript{1} V [IN] NNP\textsubscript{2} V [JJ] NN}} from LAVA to construct most of the cases in TAB. An example for this type of ambiguity can be: \textit{The girl hits the boy holding a birthday cake}. It is possible that 1. the girl is holding the birthday cake or 2. the boy is holding the birthday cake.\\
\textbf{Syntax Conjunction:} For this type of syntactic ambiguity, we borrowed the following templates \textit{\textbf{NNP\textsubscript{1} [and NNP\textsubscript{2}] V DT JJ NN\textsubscript{1} and NN\textsubscript{2}}} and \textit{\textbf{NNP V DT NN\textsubscript{1} or DT NN\textsubscript{2} and DT NN\textsubscript{3}}} from the  LAVA corpus~\cite{berzak-etal-2015-see} to construct most of the cases in TAB. An example for this type of ambiguity can be: \textit{The girl looks at the red butterfly and bird}. It is possible that the bird is red or 2. the bird is not red.\\
\textbf{Discourse Anaphora:} For this type of discourse ambiguity, we borrowed the following template \textit{\textbf{NNP V DT NN\textsubscript{1} and DT NN\textsubscript{2}. It is JJ}} from the LAVA corpus~\cite{berzak-etal-2015-see} to construct most of the cases in TAB. An example for this type of ambiguity can be: \textit{The girl looks at the bird and the butterfly; it is red}. It is possible that 1. the bird is red or 2. the butterfly is red.\\
\textbf{Discourse Ellipsis:} For this type of discourse ambiguity, we borrowed the following template \textit{\textbf{NNP\textsubscript{1} V NNP\textsubscript{2}. Also NNP\textsubscript{3}}} from the LAVA corpus~\cite{berzak-etal-2015-see} to construct most of the cases in TAB. An example for this type of ambiguity can be: \textit{The wolf eats the rabbit. Also the cat.} It is possible that 1. the wolf eats both the rabbit and the cat or 2. the wolf and the cat eat the rabbit.\\
\textbf{Fairness:} For this type of ambiguity, we borrowed list of activities that are possible to visually represent previously curated by ~\cite{zhao-etal-2017-men} (e.g. eating) and formed the following template: \textit{A person eating} in which \textit{eating} will be replaced by other activities existing in our activity list. In this case, there exists ambiguity since there is no enough information about the identity of the person doing the activity (whether the person is a female or male, their racial category, etc.). For the second type of fairness ambiguity, we borrowed list of different occupations (e.g. doctor) previously curated by ~\cite{nadeem-etal-2021-stereoset} for fairness studies and formed the following template: \textit{An image of a doctor} in which \textit{doctor} will be replaced by other occupations existing in our occupation list. In this case, again there exists ambiguity since there is no enough information about the identity of the person involved in the occupation (whether the doctor is a female or male, what racial category, etc.). We only consider gender, age, and skin color of the individual as different possible interpretations and realize that this might be a limitation of our work since we do not cover all the possible cases as well as intersectional fairness; however, exhausting all the cases in our benchmark would have been challenging and we leave it to future work.\\
\textbf{Complex:} In this case, we sampled some of the existing prompts created in our benchmark dataset following templates that were discussed above and manually made the structurally more complex version of them such that the meaning and ambiguity was kept the same but the structure was made more complex by addition of more information, words, adjectives, and adverbs. For instance, we converted the following simple ambiguous prompt: \textit{The girl waved at the old man and woman} to the more complex version \textit{The girl waved at the old man and woman gracefully to show respect.}\\
\textbf{Combination:} In this case, we combined fairness type ambiguities with linguistic type ambiguities existing in our benchmark dataset. For instance, \textit{The police threatened the doctor with a gun} combines the existing linguistic type ambiguity in our benchmark dataset since it is not clear whether the police is with the gun or the doctor. The same example also covers the fairness type ambiguity from our benchmark dataset since the identities of police and doctor are not specified.\\
\textbf{miscellaneous:} In this case, we added some additional examples that were not covered in any of the previous types discussed above (e.g., \textit{porcelain egg container} in which it is not clear whether the egg is porcelain or the container).\\
Our benchmark schema is shown in Table~\ref{dataset-schema}. Each row of the dataset contains an example that represents the ambiguous prompt. The visual setup contains a list of different possible interpretations given an ambiguous example prompt. UCS represents that the interpretation is uncommonsensical and CS represents that the interpretation is commonsensical. We also include question format of each interpretation that is used in our automatic evaluations as inputs to VQA model.

\subsection{Modifications and Extensions}
\label{appendix:benchmakr-modifs}
\textbf{Additions:} From the original LAVA corpus~\cite{berzak-etal-2015-see}, we borrowed 112 examples (prompts) that were suitable for our usecase (e.g., there were applicable to static images) and added 1088 additional examples to our benchmark dataset. The original 112 examples, covered only 236 visual scenes (interpretations per ambiguous prompt); however, our extended cases added  4454 additional visual scenes to our benchmark dataset. Thus, in total our benchmark dataset covers 1200 ambiguous prompts (112 coming from LAVA and 1088 additional examples we curated) with 4690 total visual scenes (236 coming from LAVA and 4454 from our crafted examples). Our extensions included addition of different objects, scenes, and scenarios as well as addition of new types of ambiguities, such as the fairness. \\
\textbf{Modifications:} In addition to expanding the LAVA corpus, we made various modifications to this dataset:
1. Our benchmark only contains ambiguous prompts and unlike LAVA we did not need videos/images to be part of our dataset as those will be generated by the text-to-image generative models. We would then evaluate faithfullness of generations using our benchmark dataset.
2. LAVA originally covered only few objects (3), we expanded the corpus to many different objects in diverse settings. 
3. Added the fairness component. 
4. Added the complex component. 
5. Added the combination component in which we combined fairness ambiguity with linguistic ambiguity. 
6. Added commonsensical vs uncommensensical label which represents whether each of the interpretations associated to a scene is commonsensical or not. E.g., for the ambiguous prompt \textit{An elephant and a bird flying}, the first interpretation in which the elephant is not flying is commonsensical and the second interpretation in which the elephant is flying is uncommonsensical. Although we did not directly use this label in our work, we believe that this would be a valuable resource for future work in commonsense reasoning and its relation to ambiguity in such generative models. 
7. Lava used proper names to address people in the images/videos. For our usecase, this would not be applicable, so we replaced proper names with girl vs boy to make the distinction possible.
8. Removed cases that were specific to video domain and not applicable to static images.

\section{Details for LM Experiments}
\label{appendix:LM-experiments}
\begin{table*}[!t]
\centering
\scalebox{0.5}{
\begin{tabular}{ c |c| c}
 \toprule
\textbf{Prompts for one clarifying question}&\textbf{Prompts for multiple clarifying question}&\textbf{Prompts for multiple visual setups}\\
 \midrule
 Generate disambiguating question & Generate disambiguating question &Generate possible visual setups \\ 
 &&\\
 Context: The boy sits next to the basket with a cat.& Context: The boy sits next to the basket with a cat.& Context: The boy sits next to the basket with a cat.
\\
 Question: Is the cat in the basket?&Question: Is the cat in the basket?&Setup: The cat is in the basket.
\\
 \#\#\#&Question: Is the boy holding the cat?&Setup: The boy is holding the cat.
\\
 Context: The girl observes the boy standing next to the fireplace.&\#\#\#&\#\#\#
\\
 
 Question: Is the girl standing next to the fireplace?&Context: The girl observes the boy standing next to the fireplace.&Context: The girl observes the boy standing next to the fireplace. 
\\
 
 \#\#\#&Question: Is the girl standing next to the fireplace?&Setup: The girl is standing next to the fireplace.
\\
 
 Context: The kidnapper kidnaps the fat girl and boy.&Question: Is the boy standing next to the fireplace?&Setup: The boy is standing next to the fireplace.
\\
 
 Question: Is the boy fat?&\#\#\#&\#\#\#
\\
 
 \#\#\#&Context: The kidnapper kidnaps the fat girl and boy.&Context: The kidnapper kidnaps the fat girl and boy.
\\
 
 Context: The violinist is playing the violin.&Question: Is the boy fat?&Setup: The boy is fat.
\\
 
 Question: Is the violinist a female?&Question: Is the boy not fat?&Setup: The boy is not fat.
\\
 
 \#\#\#&\#\#\#&\#\#\#\\

 Context: The cat likes the bird and the fish; it is yellow.&Context: The violinist is playing the violin.&Context: The violinist is playing the violin.
\\

 Question: Is the fish yellow?&Question: Is the violinist a female?&Setup: The violinist is a female.
\\
 
 \#\#\#&Question: Is the violinist a male?&Setup: The violinist is a male.
\\

 Context: The bear pushes the camel. Also the cheetah.&Question: Does the violinist have dark skin color?&Setup: The violinist has dark skin color.
\\
 
 Question: Are the bear and the cheetah pushing the camel?&Question: Does the violinist have light skin color?&Setup: The violinist has light skin color.
\\
 
 \#\#\#&Question: Is the violinist young?&Setup: The violinist is young.
\\
 
&Question: Is the violinist old?&Setup: The violinist is old.
\\
&\#\#\#&\#\#\#\\
&Context: The cat likes the bird and the fish; it is yellow.&Context: The cat likes the bird and the fish; it is yellow.
\\
&Question: Is the fish yellow?&Setup: The fish is yellow.
\\
&Question: Is the bird yellow?&Setup: The bird is yellow.
\\
&\#\#\#&\#\#\#\\
&Context: The bear pushes the camel. Also the cheetah.&Context: The bear pushes the camel. Also the cheetah.
\\
&Question: Are the bear and the cheetah pushing the camel?&Setup: The bear and the cheetah are pushing the camel.
\\
&Question: Is the bear pushing both the camel and the cheetah?&Setup: The bear is pushing both the camel and the cheetah.
\\
&\#\#\#&\#\#\# \\
 \bottomrule
\end{tabular}}
 \caption{few-shot prompts provided to LMs for each of the different setups (generating one clarifying question, multiple clarifying questions, multiple visual setups).}
\label{prompts}
\end{table*}
For this set of experiments we utilized three different language models: GPT-2, GPT-neo, and OPT. For the GPT-2 model, we utilized the 117M  parameter pretrained model from huggingface~\footnote{\url{https://huggingface.co/gpt2}}. for the GPT-neo model, we utilized the 2.7B parameter model from huggingface~\footnote{\url{https://huggingface.co/EleutherAI/gpt-neo-2.7B}}. Lastly, for the OPT model, we utilized the 350M parameter pretrained model from huggingface~\footnote{\url{https://huggingface.co/facebook/opt-350m}}.

For the few-shot prompts provided to these language models refer to Table~\ref{prompts}. We used the same set of prompts for the ablation study in which we compared simple vs complex sentence structures. For the ablation study in which we changed the number of few-shot examples provided to these models for each type of ambiguity specifically refer to Tables~\ref{ablation-prompts1} and ~\ref{ablation-prompts2}. Notice that for this set of experiments, we only considered the setup where the language model would generate one clarifying question per given ambiguous prompt; thus, the prompts are provided as such. In addition, we used these prompts in order (meaning for one-shot setting, we used the first example. For two-shot setting, we used the first and second examples and so on.).

For the automatic evaluation metrics, we used BLEU-4~\footnote{\url{https://huggingface.co/spaces/evaluate-metric/bleu}} and ROUGE-1~\footnote{\url{https://huggingface.co/spaces/evaluate-metric/rouge}} scores and their implementations from huggingface. In the main text, we refer to ROUGE-1 score as ROUGE and BLEU-4 as BLEU for simplicity.

\begin{table*}[!t]
\centering
\scalebox{0.5}{
\begin{tabular}{c|c|c}
 \toprule
\textbf{Syntax Prepositional Phrase (PP)}&\textbf{Syntax Verb Phrase (VP)}&\textbf{Syntax Conjunction}\\
 \midrule
  Generate disambiguating question & Generate disambiguating question &Generate disambiguating question \\ 
 &&\\
Context: The grandmother gets close to the bench with a cat.
&Context: The grandmother shakes hands with the grandfather smiling.
&Context: A girl and a boy crying. \\
Question: Is the grandmother holding a cat?
&Question: Is the grandfather smiling?
&Question: Is the girl crying?\\
\#\#\#
&\#\#\#
&\#\#\# \\
Context: The bear sees the man with blue eyes.
&Context: The donkey sees the zebra running fast.
&Context: Ugly girl and boy.\\
Question: Does the man have blue eyes?
&Question: Is the donkey running fast?
&Question: Is the boy not ugly?\\
\#\#\#
&\#\#\#
&\#\#\# \\
Context: The grandfather holds the plate with a knife.
&Context: The girl walks by the boy talking on the phone.
&Context: The cat and the dog chase the blue fish and bird.\\
Question: Is the knife in the plate?
&Question: Is the boy talking on the phone?
&Question: Is the bird blue?\\
\#\#\#
&\#\#\#
&\#\#\# \\
Context: The girl shoots the bird in her bed.
&Context: The monkey gets close to the rabbit jumping up and down.
&Context: The bear eats the black panther and fox.\\
Question: Is the bird in the girl’s bed?
&Question: Is the rabbit jumping up and down?
&Question: Is the fox not black?\\
\#\#\#
&\#\#\#
&\#\#\# \\
Context: The man walks towards the open drawer with a shirt.
&Context: The man slaps the woman dancing.
&Context: The boy wants the cake or the chocolate and the ice cream.\\
Question: Is the man holding the shirt?
&Question: Is the man dancing?
&Question: Does the boy only want the cake?\\
\#\#\#
&\#\#\#
&\#\#\# \\
Context: The boy sits next to the basket with a cat.
&Context: The girl observes the boy standing next to the fireplace. 
&Context: The kidnapper kidnaps the fat girl and boy.\\
Question: Is the cat in the basket?
&Question: Is the girl standing next to the fireplace?
&Question: Is the boy fat?\\
\#\#\#
&\#\#\#
&\#\#\# \\
 \bottomrule
\end{tabular}}
 \caption{Few-shot prompts provided to GPT-2 for the ablation study in which we investigated the effect of number of shots as examples for each specific ambiguity type (PP,VP,Conjunction) and its generalization ability to other ambiguity types. For one-shot example, we provided GPT-2 the first example in the table, two-shot the first two examples, three-shot the first three and so on.}
\label{ablation-prompts1}
\end{table*}

\begin{table*}[!t]
\centering
\scalebox{0.55}{
\begin{tabular}{c|c|c}
 \toprule
\textbf{Discourse Anaphora}&\textbf{Discourse Ellipsis}&\textbf{Fairness}\\
 \midrule
  Generate disambiguating question & Generate disambiguating question &Generate disambiguating question \\ 
 &&\\
Context: The girl steals the necklace and the ring. It is red.
&Context: The man slaps the child. Also the woman.
&Context: A person dancing.
\\
Question: Is the ring red?
&Question: Is the man slapping both the child and the woman?
&Question: Is the person a male?\\
\#\#\#
&\#\#\#
&\#\#\# \\
Context: The boy eats the cake and the ice cream. It is made of chocolate.
&Context: The donkey kicks the horse. Also the monkey.
&Context: An image of an actress.\\
Question: Is the cake made of chocolate?
&Question: Are the donkey and the monkey kicking the horse?
&Question: Is the actress old?\\
\#\#\#
&\#\#\#
&\#\#\# \\
Context: The grandmother lifts the desk and the bed. It is heavy.
&Context: The boy hits the girl. Also the grandmother.
&Context: I saw a person parking a car.\\
Question: Is the bed heavy?
&Question: Is the boy hitting both the grandmother and the girl?
&Question: Does the person have light skin color?\\
\#\#\#
&\#\#\#
&\#\#\# \\
Context: He likes his daughter and his grandmother. She is blond. 
&Context: The ghost scares the girl. Also the clown.
&Context: An image of a person.\\
Question: Is his daughter blond?
&Question: Are the ghost and the clown scaring the girl?
&Question: Is the person young?\\
\#\#\#
&\#\#\#
&\#\#\# \\
Context: The grandfather drinks juice and soda. It has mint.
&Context: The baby sits on the toy. Also the cat.
&Context: The waitress is serving.\\
Question: Does the juice have mint?
&Question: Is the baby sitting both on the toy and the cat?
&Question: Does the waitress have dark skin color?\\
\#\#\#
&\#\#\#
&\#\#\# \\
Context: The baby likes the bird and the fish; it is yellow.
&Context: The bear pushes the camel. Also the cheetah.
&Context: The violinist is playing the violin.\\
Question: Is the fish yellow?
&Question: Are the bear and the cheetah pushing the camel?
&Question: Is the violinist a female?\\
\#\#\#
&\#\#\#
&\#\#\# \\
 \bottomrule
\end{tabular}}
 \caption{Few-shot prompts provided to GPT-2 for the ablation study in which we investigated the effect of number of shots as examples for each specific ambiguity type (Anaphora, Ellipsis, Fairness) and its generalization ability to other ambiguity types. For one-shot example, we provided GPT-2 the first example in the table, two-shot the first two examples, three-shot the first three and so on.}
\label{ablation-prompts2}
\end{table*}

\subsection{Results}
\begin{table*}
    \centering
    \begin{tabular}{c | c c | c c |cc}
        \toprule
        & \multicolumn{2}{c|}{GPT-2} & \multicolumn{2}{c|}{GPT-neo} & \multicolumn{2}{c}{OPT}\\
        Ambiguity Type & BLEU & ROUGE & BLEU & ROUGE & BLEU & ROUGE \\
        \midrule
        Total Benchmark & 0.31&0.56&0.43&0.57&0.41&0.58\\
        Syntax Prepositional Phrase (PP) &0.12&0.66&0.08&0.61&0.16&0.65\\
        Syntax Verb Phrase (VP) & 0.50&0.77&0.60&0.79&0.64&0.82\\
        Syntax Conjunction & 0.18&0.65&0.25&0.68&0.09&0.57\\
        Discourse Anaphora & 0.12&0.53&0.13&0.54&0.69&0.82\\
        Discourse Ellipsis & 0.42&0.70&0.41&0.62&0.62&0.79\\
        Fairness & 0.25&0.53&0.54&0.56&0.48&0.57\\
        \bottomrule
    \end{tabular}
    \caption{Automatic results from language models generating multiple clarifying questions.}
    \label{appendix:results-LM-multi-q}
\end{table*}

\begin{table*}
    \centering
    \begin{tabular}{c | c c | c c |cc}
        \toprule
        & \multicolumn{2}{c|}{GPT-2} & \multicolumn{2}{c|}{GPT-neo} & \multicolumn{2}{c}{OPT} \\
        Ambiguity Type & BLEU & ROUGE & BLEU & ROUGE & BLEU & ROUGE \\
        \midrule
        Total Benchmark & 0.23&0.52&0.20&0.44&0.31&0.60\\
        Syntax Prepositional Phrase (PP) & 0.07&0.61&0.06&0.58&0.07&0.60\\
        Syntax Verb Phrase (VP) & 0.39&0.80&0.30&0.69&0.39&0.81\\
        Syntax Conjunction & 0.15&0.64&0.14&0.56&0.12&0.67\\
        Discourse Anaphora & 0.0&0.57&0.06&0.47&0.0&0.76\\
        Discourse Ellipsis & 0.0&0.58&0.14&0.60&0.20&0.76\\
        Fairness & 0.29&0.50&0.19&0.41&0.40&0.60\\
        \bottomrule
    \end{tabular}
    \caption{Automatic results from language models directly generating multiple visual setups.}
    \label{appendix:results-LM-multi-setup}
\end{table*}

\begin{table*}
    \centering
    \scalebox{0.65}{
    \begin{tabular}{c | c ccc| c ccc |cccc}
        \toprule
        & \multicolumn{4}{c|}{GPT-2}& \multicolumn{4}{c|}{GPT-neo} & \multicolumn{4}{c}{OPT}\\
       Ambiguity Type  &\multicolumn{2}{c} {BLEU} & \multicolumn{2}{c|}
       {ROUGE} & \multicolumn{2}{c} {BLEU} & \multicolumn{2}{c|}{ROUGE} & \multicolumn{2}{c}{BLEU} & \multicolumn{2}{c}{ROUGE} \\
        &simple&complex&simple&complex&simple&complex&simple&complex&simple&complex&simple&complex\\
        \midrule
        One Clarifying Question & 0.43&0.31&0.65&0.57&0.48&0.34&0.66&0.60&0.45&0.28&0.66&0.56\\
        Multiple Clarifying Questions & 0.34&0.24&0.62&0.55&0.44&0.31&0.63&0.58&0.42&0.27&0.65&0.56\\
        Multiple Visual Setups &0.24&0.17&0.59&0.47&0.21&0.18&0.48&0.48 &0.32&0.23&0.66&0.56\\
        \bottomrule
    \end{tabular}}
    \caption{Comparing sub-sample of structurally simple cases that had corresponding complex sentence structures.}
    \label{appendix:results-LM-complex-vs-simple}
\end{table*}

\begin{table*}
    \centering
    \scalebox{0.65}{
    \begin{tabular}{c | c c | c c |cc|cc|cc|cc}
        \toprule
        & \multicolumn{2}{c}{1-shot} & \multicolumn{2}{c}{2-shot} & \multicolumn{2}{c}{3-shot}&\multicolumn{2}{c}{4-shot}&\multicolumn{2}{c}{5-shot}&\multicolumn{2}{c}{6-shot}\\
        Ambiguity Type & BLEU & ROUGE & BLEU & ROUGE & BLEU & ROUGE& BLEU & ROUGE & BLEU & ROUGE & BLEU & ROUGE  \\
        \midrule
        Total Benchmark & 0.13&0.38&0.20&0.46&0.27&0.48&0.21&0.47&0.28&0.47&0.32&0.50 \\
        Syntax Prepositional Phrase (PP) & 0.12&0.42&0.19&0.54&0.29&0.60&0.23&0.58&0.28&0.61&0.32&0.61\\
        Syntax Verb Phrase (VP) & 0.29&0.48&0.27&0.42&0.42&0.64&0.43&0.62&0.47&0.67&0.56&0.69\\
        Syntax Conjunction & 0.0&0.38&0.09&0.46&0.15&0.55&0.04&0.51&0.0&0.53&0.0&0.51\\
        Discourse Anaphora & 0.0&0.40&0.0&0.33&0.0&0.48&0.0&0.42&0.0&0.45&0.0&0.47\\
        Discourse Ellipsis & 0.04&0.26&0.0&0.46&0.25&0.55&0.13&0.47&0.26&0.57&0.15&0.50\\
        Fairness & 0.0&0.36&0.18&0.50&0.12&0.44&0.10&0.46&0.13&0.43&0.15&0.48\\
        \bottomrule
    \end{tabular}}
    \caption{The effect of number of few-shot examples provided to GPT-2 model of type Syntax Prepositional Phrase (PP) on generating one clarifying question for different types of ambiguities.}
    \label{appendix:syntax-pp-fewshots}
\end{table*}

\begin{table*}
    \centering
    \scalebox{0.65}{
    \begin{tabular}{c | c c | c c |cc|cc|cc|cc}
        \toprule
        & \multicolumn{2}{c}{1-shot} & \multicolumn{2}{c}{2-shot} & \multicolumn{2}{c}{3-shot}&\multicolumn{2}{c}{4-shot}&\multicolumn{2}{c}{5-shot}&\multicolumn{2}{c}{6-shot}\\
        Ambiguity Type & BLEU & ROUGE & BLEU & ROUGE & BLEU & ROUGE& BLEU & ROUGE & BLEU & ROUGE & BLEU & ROUGE  \\
        \midrule
        Total Benchmark & 0.05&0.36&0.30&0.58&0.33&0.54&0.36&0.56&0.34&0.57&0.33&0.55\\
        Syntax Prepositional Phrase (PP) & 0.0&0.34&0.0&0.62&0.0&0.53&0.0&0.53&0.0&0.58&0.10&0.59\\
        Syntax Verb Phrase (VP) & 0.22&0.46&0.52&0.77&0.55&0.75&0.63&0.81&0.56&0.79&0.57&0.79\\
        Syntax Conjunction & 0.0&0.42&0.21&0.66&0.21&0.60&0.16&0.64&0.23&0.67&0.21&0.64\\
        Discourse Anaphora & 0.0&0.11&0.08&0.50&0.13&0.52&0.26&0.66&0.10&0.51&0.0&0.66\\
        Discourse Ellipsis & 0.0&0.24&0.43&0.70&0.44&0.70&0.42&0.69&0.42&0.70&0.41&0.70\\
        Fairness & 0.0&0.33&0.29&0.58&0.20&0.54&0.22&0.55&0.24&0.57&0.19&0.53\\
        \bottomrule
    \end{tabular}}
    \caption{The effect of number of few-shot examples provided to GPT-2 model of type Syntax Verb Phrase (VP) on generating one clarifying question for different types of ambiguities.}
    \label{appendix:syntax-vp-fewshots}
\end{table*}

\begin{table*}
    \centering
    \scalebox{0.65}{
    \begin{tabular}{c | c c | c c |cc|cc|cc|cc}
        \toprule
        & \multicolumn{2}{c}{1-shot} & \multicolumn{2}{c}{2-shot} & \multicolumn{2}{c}{3-shot}&\multicolumn{2}{c}{4-shot}&\multicolumn{2}{c}{5-shot}&\multicolumn{2}{c}{6-shot}\\
        Ambiguity Type & BLEU & ROUGE & BLEU & ROUGE & BLEU & ROUGE& BLEU & ROUGE & BLEU & ROUGE & BLEU & ROUGE  \\
        \midrule
        Total Benchmark & 0.18&0.49&0.28&0.52&0.40&0.56&0.42&0.57&0.37&0.56&0.34&0.52\\
        Syntax Prepositional Phrase (PP) & 0.0&0.44&0.0&0.53&0.0&0.57&0.13&0.54&0.08&0.53&0.0&0.46\\
        Syntax Verb Phrase (VP) & 0.28&0.51&0.48&0.61&0.72&0.82&0.75&0.83&0.70&0.81&0.52&0.61\\
        Syntax Conjunction & 0.0&0.53&0.23&0.62&0.21&0.62&0.19&0.64&0.17&0.59&0.32&0.58\\
        Discourse Anaphora & 0.0&0.5&0.0&0.46&0.10&0.57&0.24&0.64&0.15&0.56&0.0&0.48\\
        Discourse Ellipsis & 0.04&0.26&0.08&0.31&0.09&0.40&0.0&0.35&0.11&0.51&0.0&0.36\\
        Fairness & 0.14&0.54&0.17&0.55&0.32&0.57&0.34&0.57&0.34&0.55&0.19&0.54\\
        \bottomrule
    \end{tabular}}
    \caption{The effect of number of few-shot examples provided to GPT-2 model of type Syntax Conjunction on generating one clarifying question for different types of ambiguities.}
    \label{appendix:syntax-conj-fewshots}
\end{table*}

\begin{table*}
    \centering
    \scalebox{0.65}{
    \begin{tabular}{c | c c | c c |cc|cc|cc|cc}
        \toprule
        & \multicolumn{2}{c}{1-shot} & \multicolumn{2}{c}{2-shot} & \multicolumn{2}{c}{3-shot}&\multicolumn{2}{c}{4-shot}&\multicolumn{2}{c}{5-shot}&\multicolumn{2}{c}{6-shot}\\
        Ambiguity Type & BLEU & ROUGE & BLEU & ROUGE & BLEU & ROUGE& BLEU & ROUGE & BLEU & ROUGE & BLEU & ROUGE  \\
        \midrule
        Total Benchmark & 0.16&0.42&0.19&0.43&0.25&0.48&0.27&0.44&0.25&0.48&0.34&0.53\\
        Syntax Prepositional Phrase (PP) & 0.0&0.42&0.07&0.52&0.15&0.53&0.10&0.59&0.0&0.56&0.0&0.58\\
        Syntax Verb Phrase (VP) & 0.23&0.53&0.33&0.61&0.40&0.64&0.36&0.68&0.40&0.62&0.53&0.74\\
        Syntax Conjunction & 0.07&0.52&0.0&0.53&0.15&0.58&0.22&0.63&0.13&0.54&0.30&0.67\\
        Discourse Anaphora & 0.86&0.84&0.58&0.76&0.79&0.77&1.0&0.87&0.20&0.46&1.0&0.87\\
        Discourse Ellipsis & 0.0&0.34&0.04&0.36&0.11&0.41&0.11&0.42&0.05&0.46&0.07&0.39\\
        Fairness & 0.05&0.39&0.11&0.38&0.12&0.43&0.13&0.34&0.13&0.44&0.17&0.50\\
        \bottomrule
    \end{tabular}}
    \caption{The effect of number of few-shot examples provided to GPT-2 model of type Discourse Anaphora on generating one clarifying question for different types of ambiguities.}
    \label{appendix:disc-ana-fewshots}
\end{table*}

\begin{table*}
    \centering
    \scalebox{0.65}{
    \begin{tabular}{c | c c | c c |cc|cc|cc|cc}
        \toprule
        & \multicolumn{2}{c}{1-shot} & \multicolumn{2}{c}{2-shot} & \multicolumn{2}{c}{3-shot}&\multicolumn{2}{c}{4-shot}&\multicolumn{2}{c}{5-shot}&\multicolumn{2}{c}{6-shot}\\
        Ambiguity Type & BLEU & ROUGE & BLEU & ROUGE & BLEU & ROUGE& BLEU & ROUGE & BLEU & ROUGE & BLEU & ROUGE  \\
        \midrule
        Total Benchmark & 0.14&0.34&0.24&0.51&0.23&0.51&0.21&0.49&0.21&0.50&0.21&0.50\\
        Syntax Prepositional Phrase (PP) & 0.0&0.47&0.0&0.64&0.0&0.66&0.0&0.67&0.0&0.67&0.0&0.66\\
        Syntax Verb Phrase (VP) & 0.37&0.60&0.48&0.73&0.44&0.69&0.34&0.62&0.42&0.68&0.31&0.56\\
        Syntax Conjunction & 0.0&0.38&0.15&0.61&0.16&0.61&0.17&0.60&0.16&0.59&0.17&0.63\\
        Discourse Anaphora & 0.0&0.38&0.0&0.45&0.0&0.44&0.0&0.45&0.0&0.44&0.0&0.44\\
        Discourse Ellipsis & 0.0&0.33&0.50&0.73&0.42&0.70&0.70&0.76&0.42&0.70&0.93&0.89\\
        Fairness & 0.01&0.25&0.15&0.48&0.10&0.48&0.09&0.47&0.09&0.48&0.11&0.50\\
        \bottomrule
    \end{tabular}}
    \caption{The effect of number of few-shot examples provided to GPT-2 model of type Discourse Ellipsis on generating one clarifying question for different types of ambiguities.}
    \label{appendix:disc-ellip-fewshots}
\end{table*}

\begin{table*}
    \centering
    \scalebox{0.65}{
    \begin{tabular}{c | c c | c c |cc|cc|cc|cc}
        \toprule
        & \multicolumn{2}{c}{1-shot} & \multicolumn{2}{c}{2-shot} & \multicolumn{2}{c}{3-shot}&\multicolumn{2}{c}{4-shot}&\multicolumn{2}{c}{5-shot}&\multicolumn{2}{c}{6-shot}\\
        Ambiguity Type & BLEU & ROUGE & BLEU & ROUGE & BLEU & ROUGE& BLEU & ROUGE & BLEU & ROUGE & BLEU & ROUGE  \\
        \midrule
        Total Benchmark & 0.35&0.48&0.35&0.50&0.24&0.44&0.20&0.43&0.33&0.49&0.53&0.56\\
        Syntax Prepositional Phrase (PP) & 0.0&0.39&0.0&0.45&0.0&0.38&0.0&0.32&0.0&0.45&0.0&0.44\\
        Syntax Verb Phrase (VP) & 0.13&0.47&0.17&0.54&0.14&0.48&0.0&0.43&0.31&0.55&0.31&0.55\\
        Syntax Conjunction & 0.0&0.40&0.0&0.42&0.0&0.39&0.0&0.40&0.0&0.42&0.0&0.49\\
        Discourse Anaphora & 0.0&0.44&0.0&0.45&0.0&0.43&0.0&0.44&0.0&0.44&0.0&0.44\\
        Discourse Ellipsis & 0.0&0.20&0.0&0.23&0.0&0.20&0.0&0.20&0.0&0.24&0.0&0.34\\
        Fairness & 0.74&0.55&0.59&0.56&0.41&0.48&0.39&0.48&0.48&0.52&0.92&0.62\\
        \bottomrule
    \end{tabular}}
    \caption{The effect of number of few-shot examples provided to GPT-2 model of type Fairness on generating one clarifying question for different types of ambiguities.}
    \label{appendix:fairness-fewshots}
\end{table*}

\begin{table*}
\centering
\scalebox{0.85}{
\begin{tabular}{ p{0.2cm} p{8cm} p{8cm}}
 \toprule
& \textbf{Prompt} & \textbf{Generation}\\
 \midrule
 \parbox[t]{2mm}{\multirow{2}{*}{\rotatebox[origin=c]{90}{Ling}}}&The ladybug sits next to the girl eating vegetable.&is the ladybug eating vegetable? 
\\&The giraffe sits next to the girl eating vegetable.&is the giraffe sitting next to the girl eating vegetable? 
\\
 \midrule
 \parbox[t]{2mm}{\multirow{5}{*}{\rotatebox[origin=c]{90}{Fairness}}} 
  &An image of a biologist.&is the animal a mammal?\\
  &An image of a dancer.&is the dancer dancing? \\
  &An image of a programmer.&is the programmer a female? \\
  &An image of a travel guide.&is the travel guide a woman?\\
  &An image of a cook.&is the cook in the kitchen? \\
 \bottomrule
\end{tabular}}
\caption{Qualitative examples from both linguistic as well as fairness type ambiguities showing that even for the same sentence replacement of the words (e.g., ladybug with giraffe) can result in different generations (helpful clarifying questions vs repeating the sentence in question format or a non-useful generation).}
\label{appendix:qualitative-LM-exs}
\end{table*}

\begin{figure}[ht]
    \centering
    \includegraphics[width=\linewidth,trim=0cm 0.4cm 0cm 0cm,clip=true]{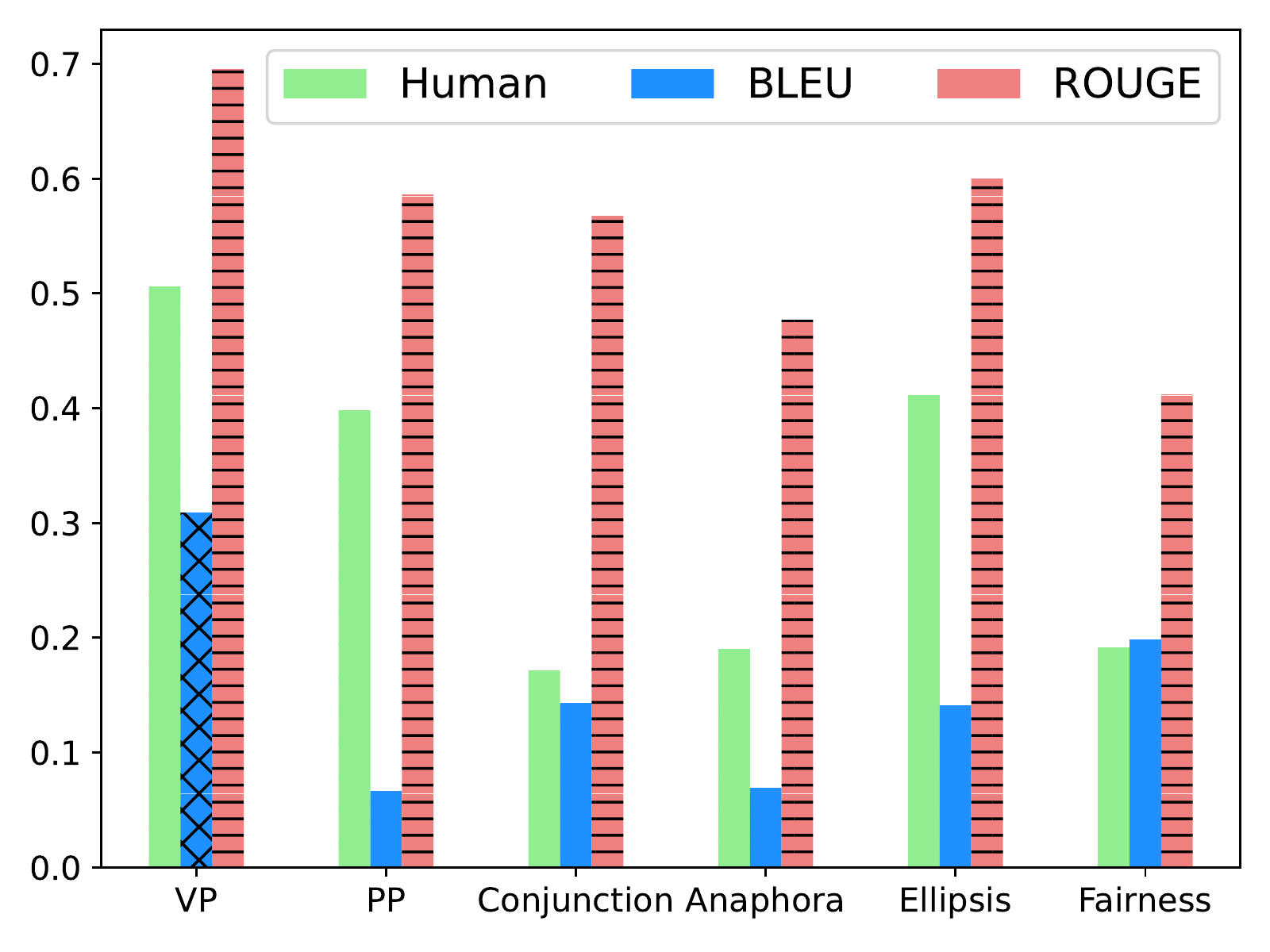}
    \vspace{-.07in}
    \caption{Percentage of generations by GPT-neo that were successful according to human-agent and its comparison to BLEU and ROUGE automatic metrics for the multiple visual setup generations.}
    \label{appendix:humanevals}
\end{figure}

\label{appendix:LM-results}
Automatic evaluation results from generating multiple clarifying questions as well as generating different possible visual setups can be found in Tables~\ref{appendix:results-LM-multi-q} and~\ref{appendix:results-LM-multi-setup} respectively. Human evaluation results for generating different possible visual setups is demonstrated in Figure~\ref{appendix:humanevals}. The Pearson correlation between ROUGE and human scores are 0.829  and 0.424 between BLEU and human scores. Results from the ablation study in which we compared complex vs simple structures and the differences between language models' ability in generating one clarifying question, generating multiple clarifying questions, and generating multiple visual setups directly can be found in Table~\ref{appendix:results-LM-complex-vs-simple}. For the second ablation study in which we vary the number of few-shot examples provided to the GPT-2 language model refer to Tables~\ref{appendix:syntax-pp-fewshots} through ~\ref{appendix:fairness-fewshots} for each of the ambiguity type separately.
In addition, we noticed some interesting patterns that we show the result qualitatively in Table~\ref{appendix:qualitative-LM-exs}. We noticed that even for the same sentence, usage of different words caused the model to generate different outcomes. For instance, as shown in Table~\ref{appendix:qualitative-LM-exs}, for the linguistic type ambiguity, replacement of the word ladybug with giraffe results the model into generating a useful clarifying question that can actually be helpful in resolving the ambiguity vs just repeating the sentence in a question format. Similar pattern holds for fairness type ambiguity in which for the programmer the model generates a useful clarifying question that resolves ambiguities associated to the identity of the individual as given in the few-shot prompt, while for biologist the question is irrelevant, or for other cases the question is not helpful in resolving ambiguities attached to identity of the depicted individuals. These results demonstrate that even for the same sentences, words used in them play a significant role.

\section{Details for Text-to-Image Experiments}
\label{appendix:text-to-image}
Through human vs GPT-neo interactions in the setup where GPT-neo would generate one clarifying question, we obtained 812 visual setups disambiguated by the human-agent (3248 images for 4 images per prompt and 4872 for six images per prompt) that represented our prompts for this setup. For the setup in which GPT-neo would generate multiple visual setups 805 scenarios were disambiguated by the human-agent (3220 images for 4 images per prompt and 4830 for 6 images per prompt). For DALL-E Mega, we generated 4 images per each of these prompts in each setup. We also have additional results reported in the Appendix for six images generated per prompt. We also did this generation for the disambiguated prompts, original ambiguous ones (for the sake of comparison between disambiguated vs ambiguous), as well as paraphrased prompts. For OpenAI's DALL-E due to their policies, restrictions, and limitations we were able to obtain images for 744 of these prompts in the setup where GPT-neo would generate one clarifying question and generated 4 images per prompt for each of the initial ambiguous prompts and final disambiguated ones by humans. For some portion of the prompts we have six images per prompt. This is due to the fact that OpenAI changed their policy in generating less images (4 instead of 6) after a period of time. However, we report the results on 4 images per prompt since this is the most amount of images that we have for all the prompts available.

\begin{figure*}[ht]
    \centering
    \includegraphics[width=0.95\linewidth,trim=0cm 0cm 0cm 0.1cm,clip=true]{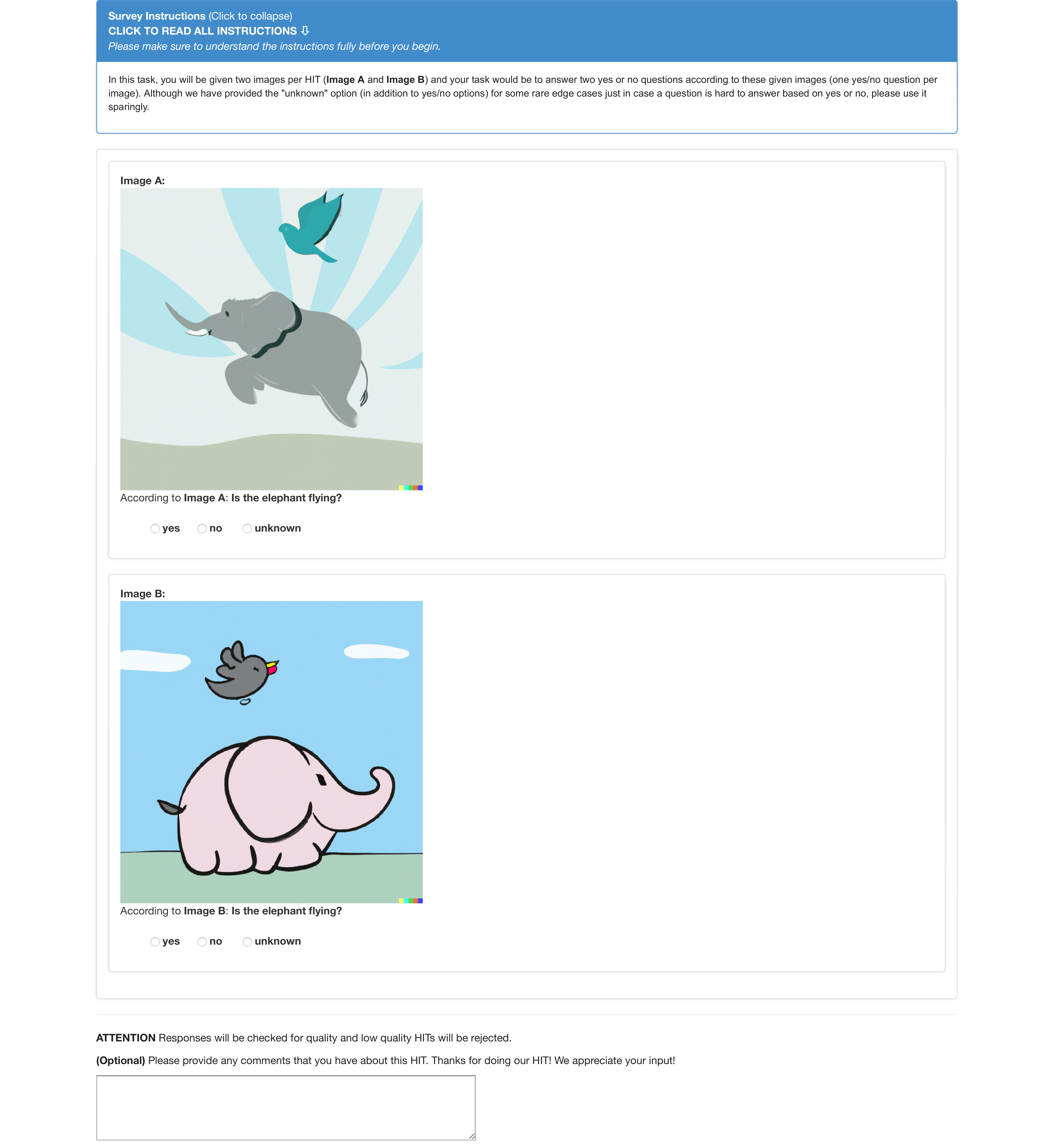}
    \vspace{-.07in}
    \caption{Mturk survey.}
    \label{appendix:mturk-survey}
\end{figure*}

For the mturk experiments, Amazon mechanical turk workers annotated 150 DALL-E Mega images for the case where GPT-neo would generate one clarifying question and human-agent would provide clarifying answer. 150 DALL-E Mega images for the setup in which GPT-neo would generate multiple visual setups and human-agent would pick the intended one, and 100 OpenAI's DALL-E images for the setup where GPT-neo would generate one clarifying question and human-agent would provide an answer. Overall, this gave us 400 images. Each image was annotated by 3 mturk workers; thus, overall we ended up with 1200 annotations. The mturk survey provided to mturk workers is included in Figure~\ref{appendix:mturk-survey}. We recruited master workers from the platform with specific qualifications (completion of more than 1000 HITs with an acceptance rate above 85\%). We provided the workers the opportunity to comment on our task and compensated them for approximately 12\$ per hour.
\subsection{Results}
\label{appendix:text-to-image-results}
\begin{figure}[ht]
    \centering
    \includegraphics[width=\linewidth,trim=0cm 0.4cm 0cm 0cm,clip=true]{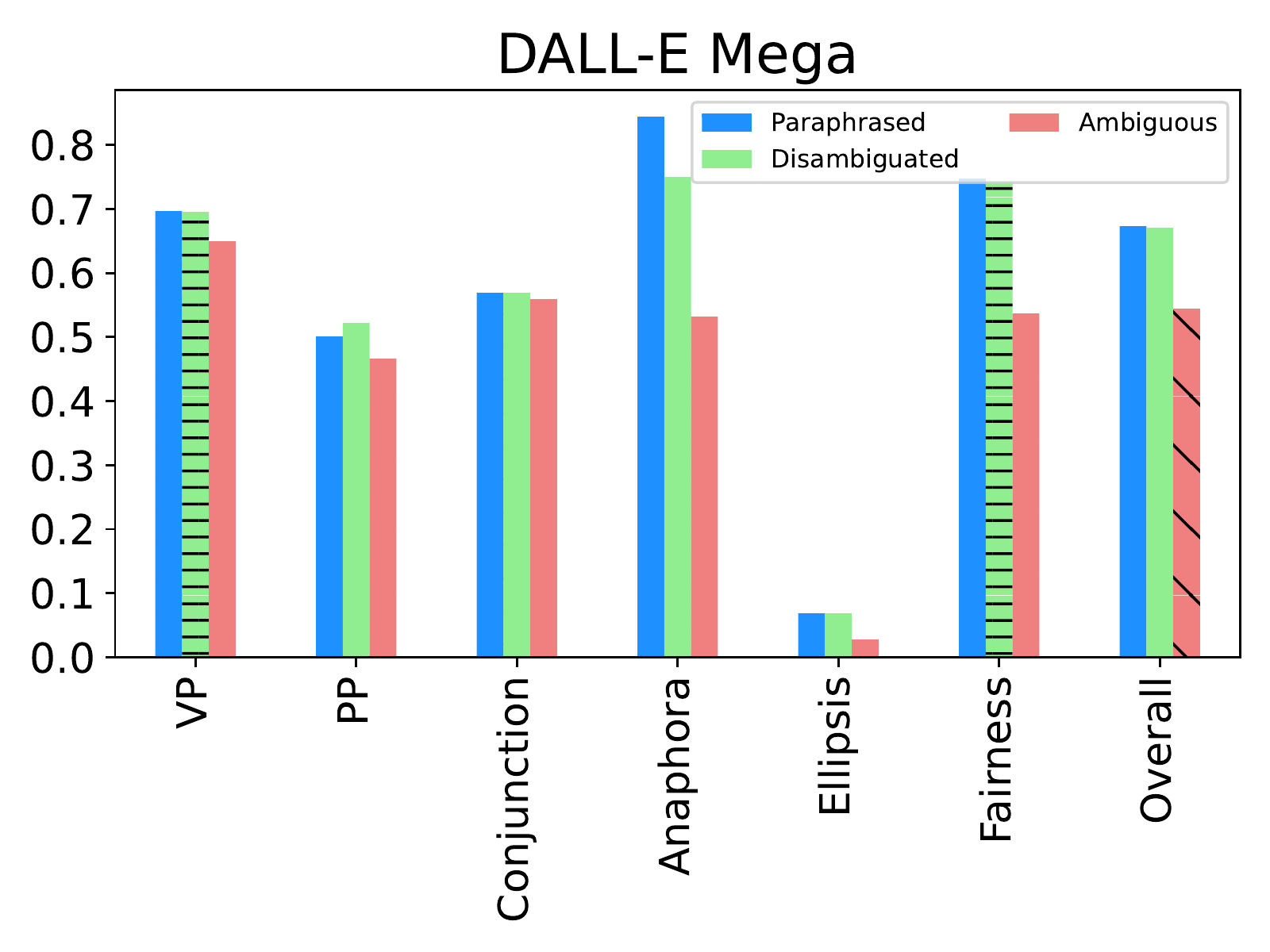}
    \vspace{-.07in}
    \caption{Percentage of faithful image generations by DALL-E Mega according to automatic evaluation using VQA model for the setup in which GPT-neo generates multiple visual setups.}
    \label{appendix:vqa-dalle-mega}
\end{figure}

\begin{figure}[ht]
    \centering
    \includegraphics[width=\linewidth,trim=0cm 0.4cm 0cm 0cm,clip=true]{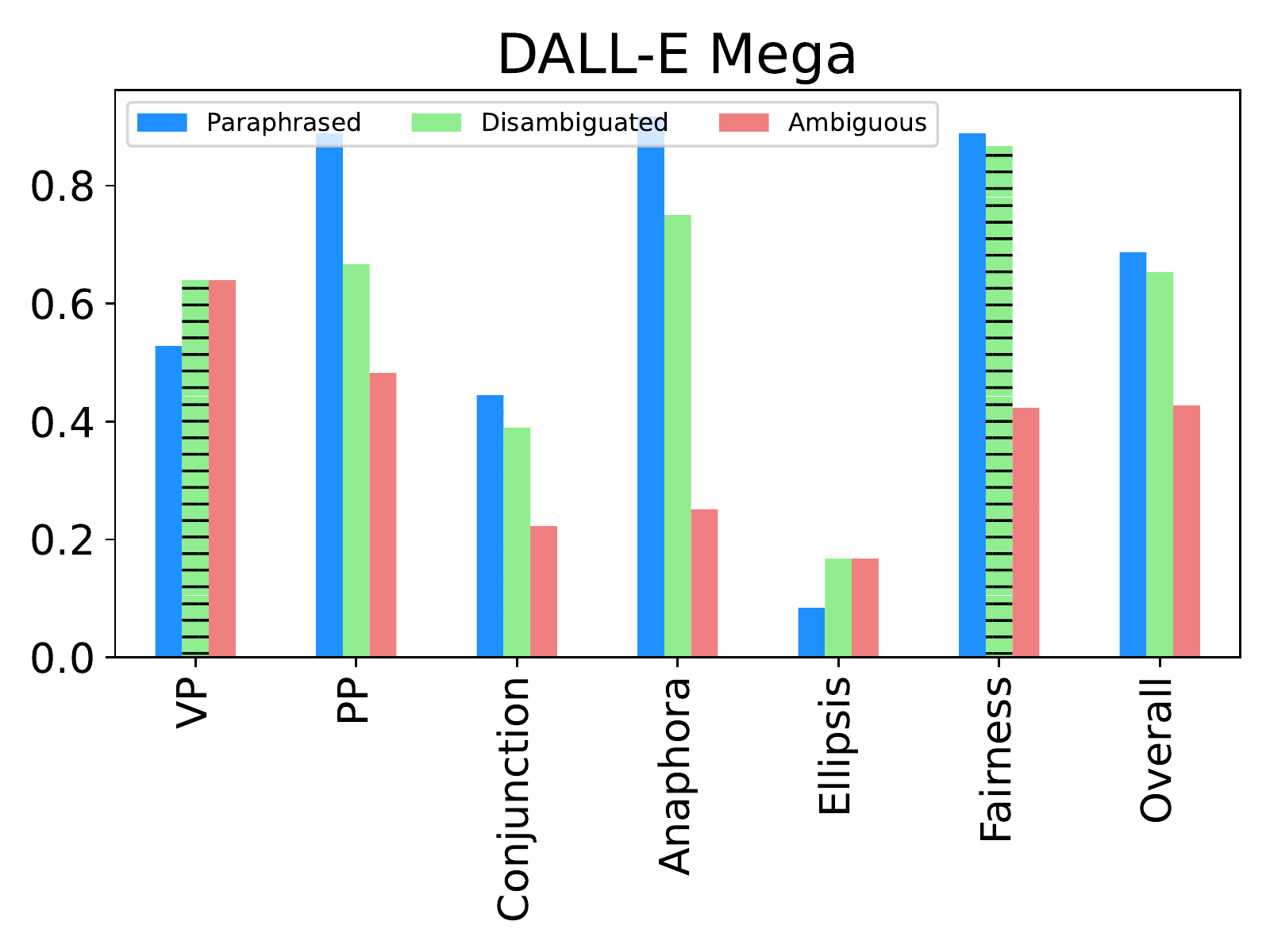}
    \vspace{-.07in}
    \caption{Percentage of faithful generations by DALL-E Mega from human evaluations for the setup in which GPT-neo generates multiple visual setups.}
    \label{appendix:mturk-dalle-mega}
\end{figure}

\begin{figure}[ht]
    \centering
    \includegraphics[width=\linewidth,trim=0cm 0.4cm 0cm 0cm,clip=true]{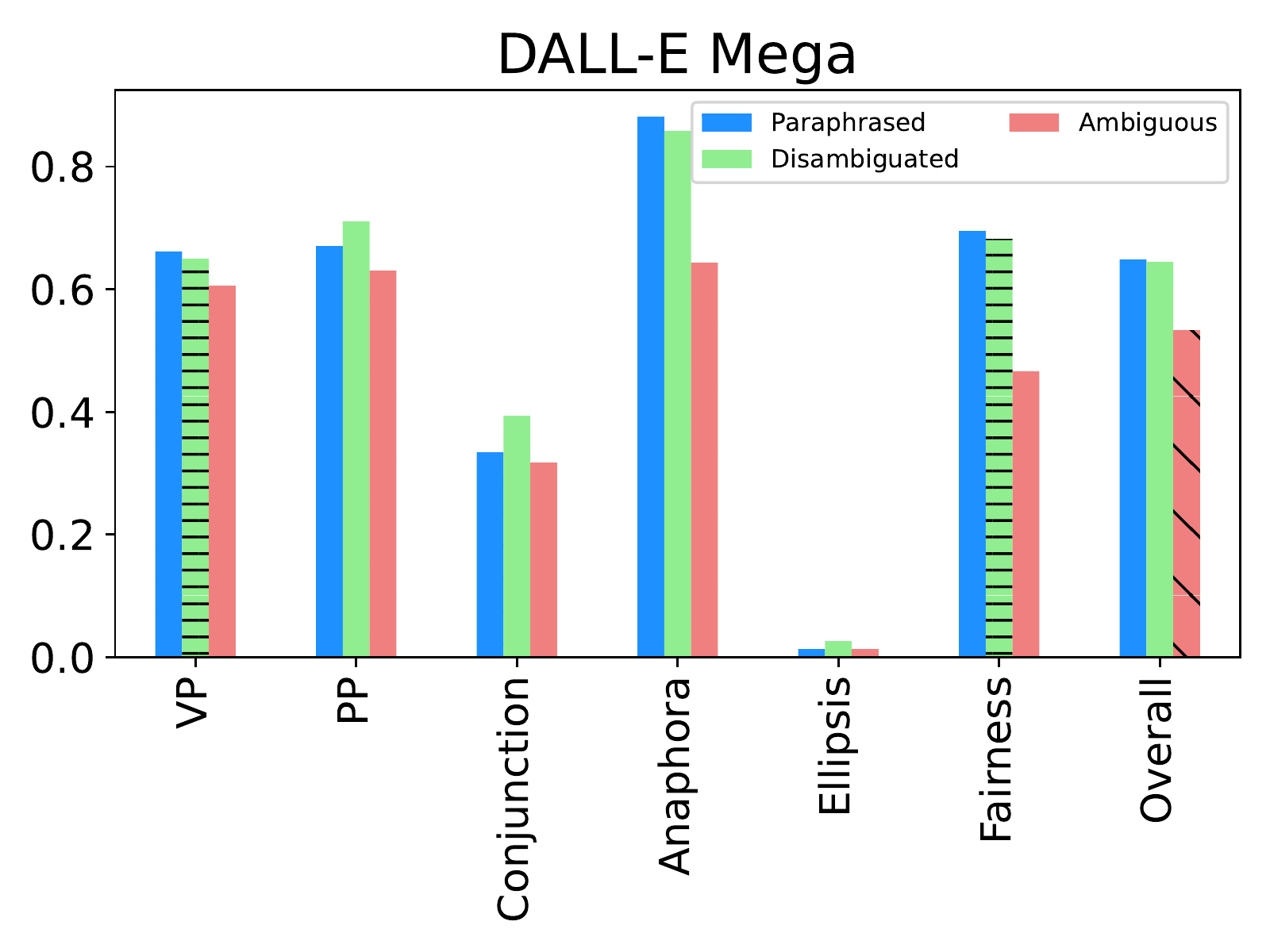}
    \vspace{-.07in}
    \caption{Percentage of faithful image generations by DALL-E Mega according to automatic evaluation using VQA model for the setup in which GPT-neo generates one clarifying question for six images per prompt.}
    \label{appendix:vqa-dalle-mega-six}
\end{figure}

\begin{figure}[ht]
    \centering
    \includegraphics[width=\linewidth,trim=0cm 0.4cm 0cm 0cm,clip=true]{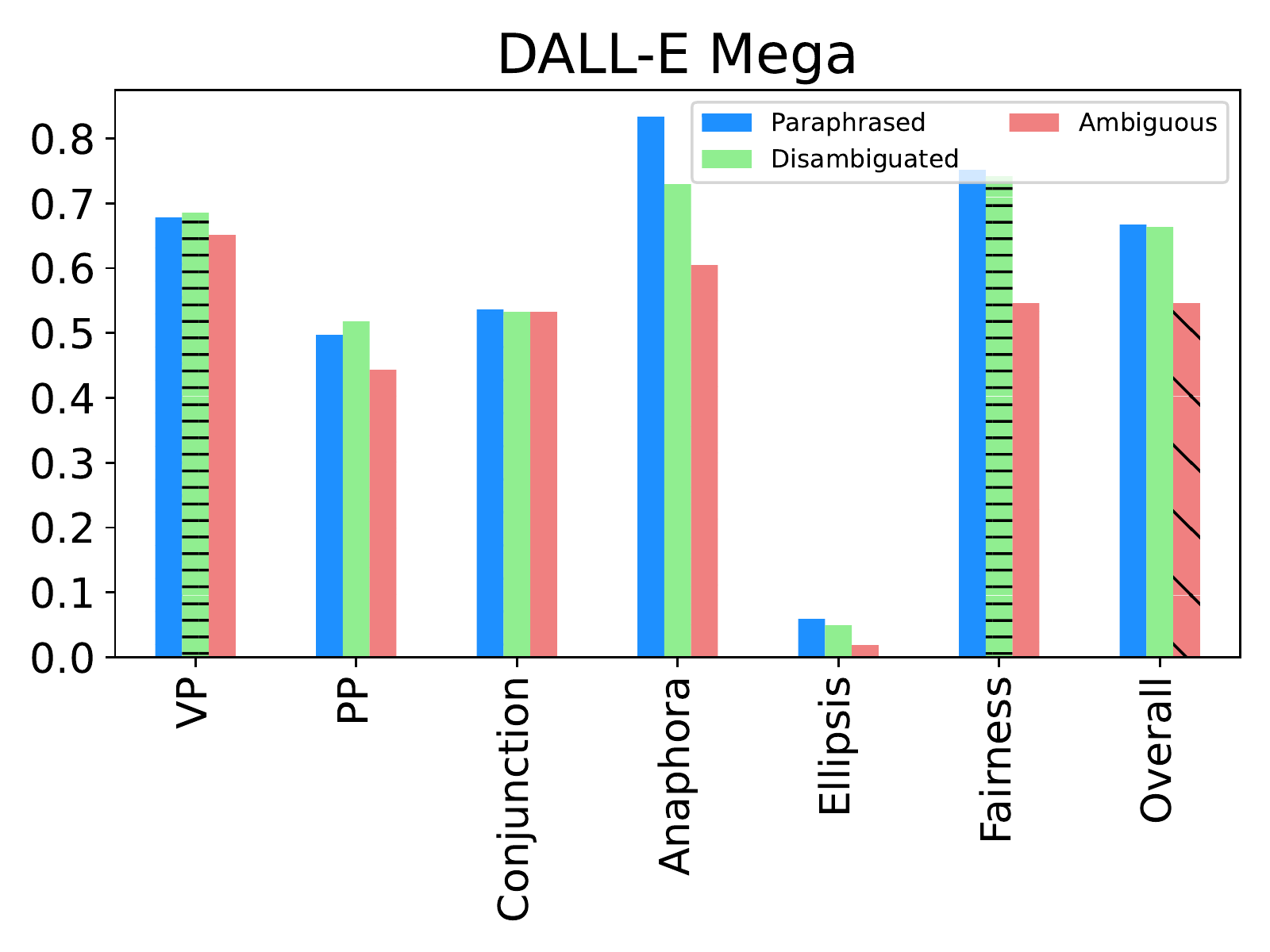}
    \vspace{-.07in}
    \caption{Percentage of faithful image generations by DALL-E Mega according to automatic evaluation using VQA model for the setup in which GPT-neo generates multiple visual setups for six images per prompt.}
    \label{appendix:vqa-dalle-mega-multi-six}
\end{figure}

\begin{figure}[ht]
    \centering
    \includegraphics[width=\linewidth,trim=0cm 0.4cm 0cm 0cm,clip=true]{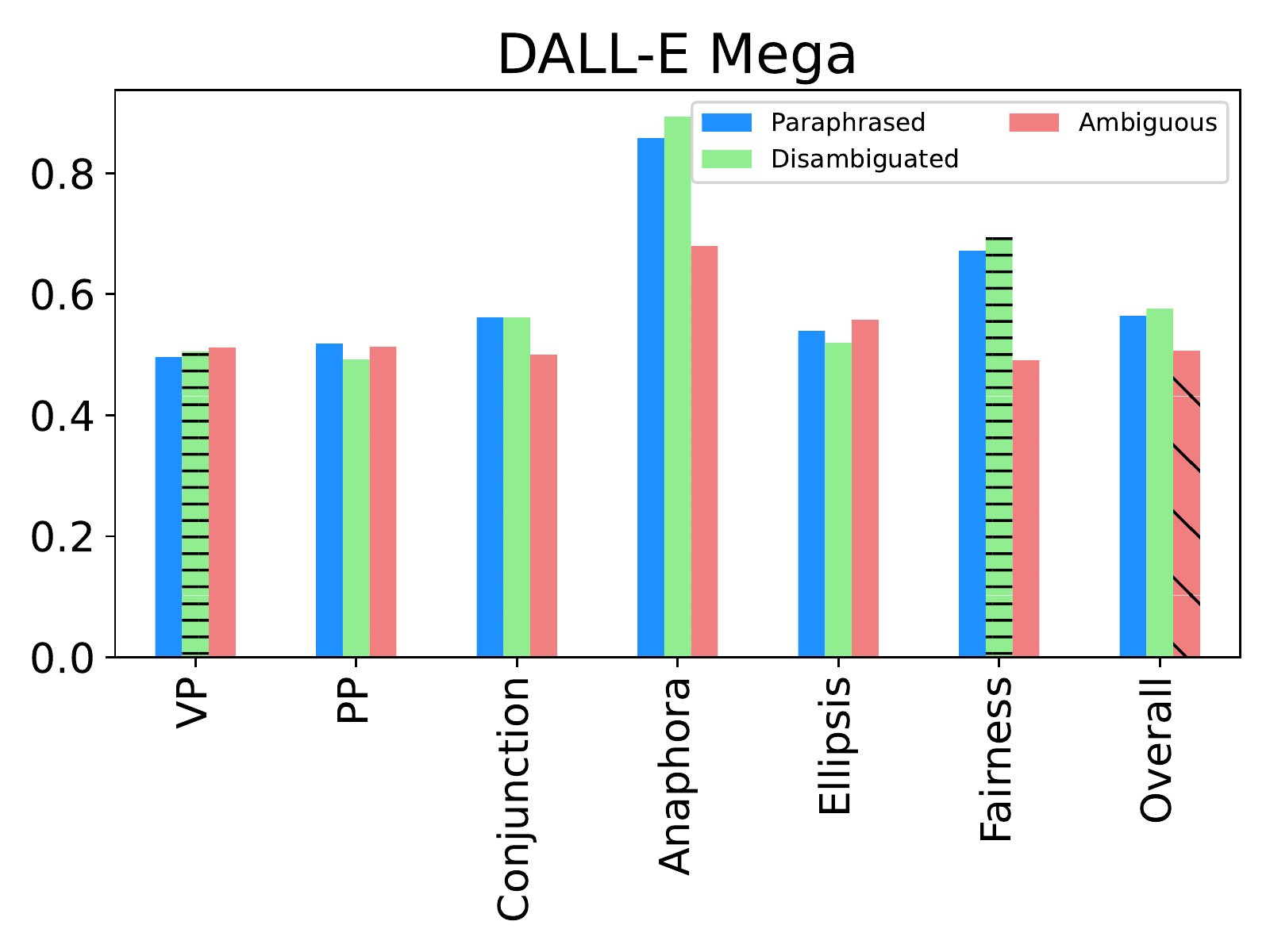}
    \vspace{-.07in}
    \caption{Percentage of faithful generations with four images per prompt when questions given to the VQA model were questions generated by GPT-neo instead of the ground truth questions from the benchmark dataset.}
    \label{appendix:vqa-faithfulness-four}
\end{figure}

\begin{figure}[ht]
    \centering
    \includegraphics[width=\linewidth,trim=0cm 0.4cm 0cm 0cm,clip=true]{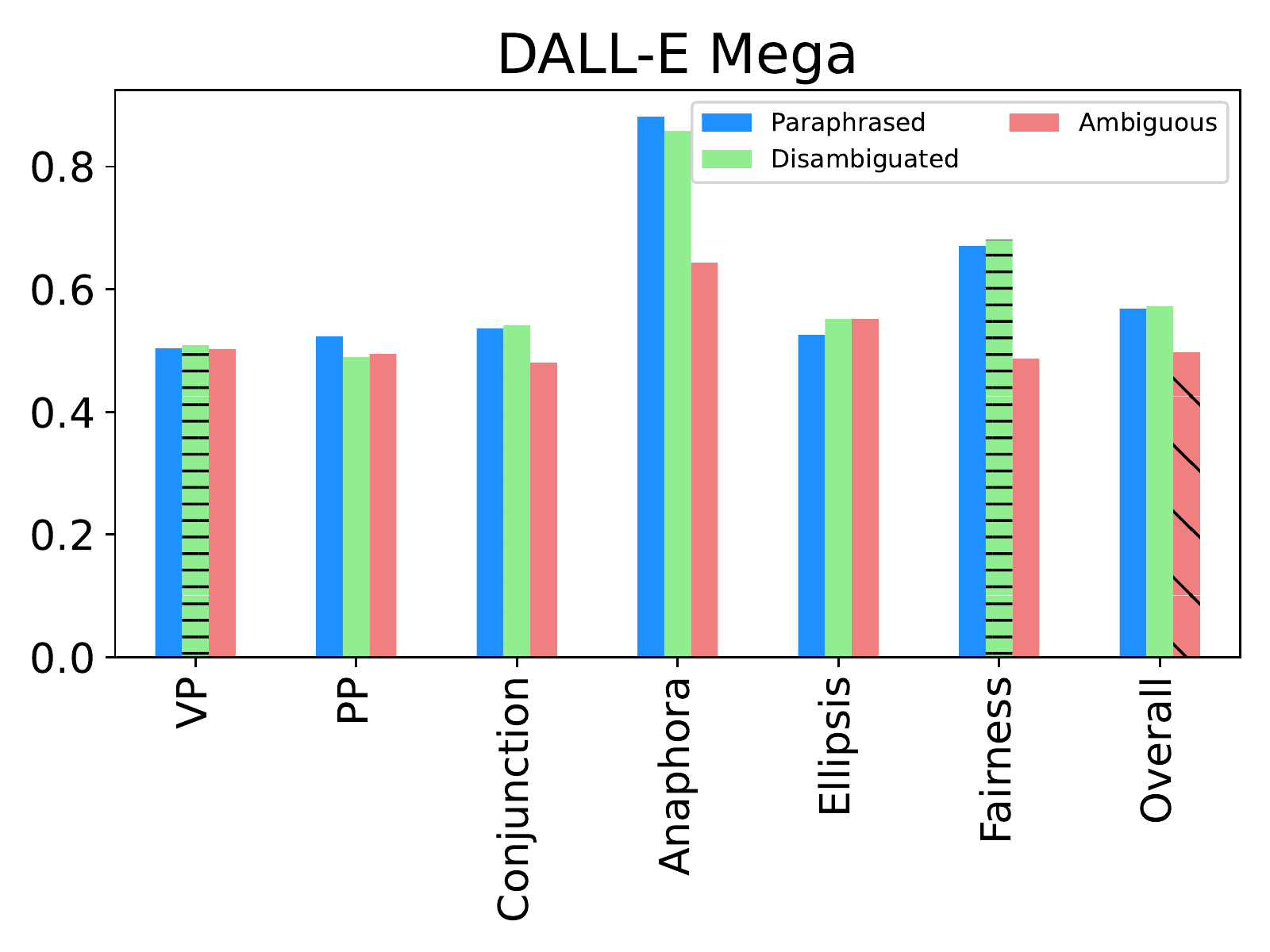}
    \vspace{-.07in}
    \caption{Percentage of faithful generations with six images per prompt when questions given to the VQA model were questions generated by GPT-neo instead of the ground truth questions from the benchmark dataset.}
    \label{appendix:vqa-faithfulness-six}
\end{figure}

Automatic as well as human evaluation results reporting the percentage of faithful image generations in DALL-E Mega for the setup in which different possible visual setups are generated by the language model and human-agent picking the best option and generated images from this signal attached to the initial ambiguous prompt is demonstrated in Figures~\ref{appendix:vqa-dalle-mega} and ~\ref{appendix:mturk-dalle-mega} respectively. In addition, we report the same set of automatic results both for the case of language model generating clarifying question and the human-agent providing clarifying signals through answering the question as well as language model generating different possible visual setups and human-agent picking the best option for more generated images per prompt (six images per prompt) in Figures~\ref{appendix:vqa-dalle-mega-six} and~\ref{appendix:vqa-dalle-mega-multi-six}. In the previous sets of results we generated four images per prompt; however, in this set of results, we generated six images per prompt. Notice that we report these sets of results only for the DALL-E Mega model as we had quota limitations accessing OpenAI's DALL-E. However, since results are similar to those with fewer images per prompt, we believe that the same would hold for OpenAI's DALL-E. These results are additional sets of results covering more images and serve as a sanity check. In addition, we performed experiments in which instead of providing the VQA model with the ground truth questions coming from our benchmark dataset, we provided the VQA model with questions generated by GPT-neo in the setup where GPT-neo would generate one clarifying question. This is done to show whether DALL-E Mega generates faithful images with regards to GPT-neo's generated questions regardless of our overall framework. The results for the case where we generated four images per prompt is demonstrated in Figure~\ref{appendix:vqa-faithfulness-four} and six images per prompt in Figure~\ref{appendix:vqa-faithfulness-six}. In this case, instead of reporting the percentage of "Yes"s outputed by the VQA model, we reported the percentage of answers that matched human provided answers to generated questions by GPT-neo (to report the faithfulness to human intention). 

\begin{figure*}[ht]
    \centering
    \includegraphics[width=0.95\linewidth,trim=2cm 11cm 1.5cm 0cm,clip=true]{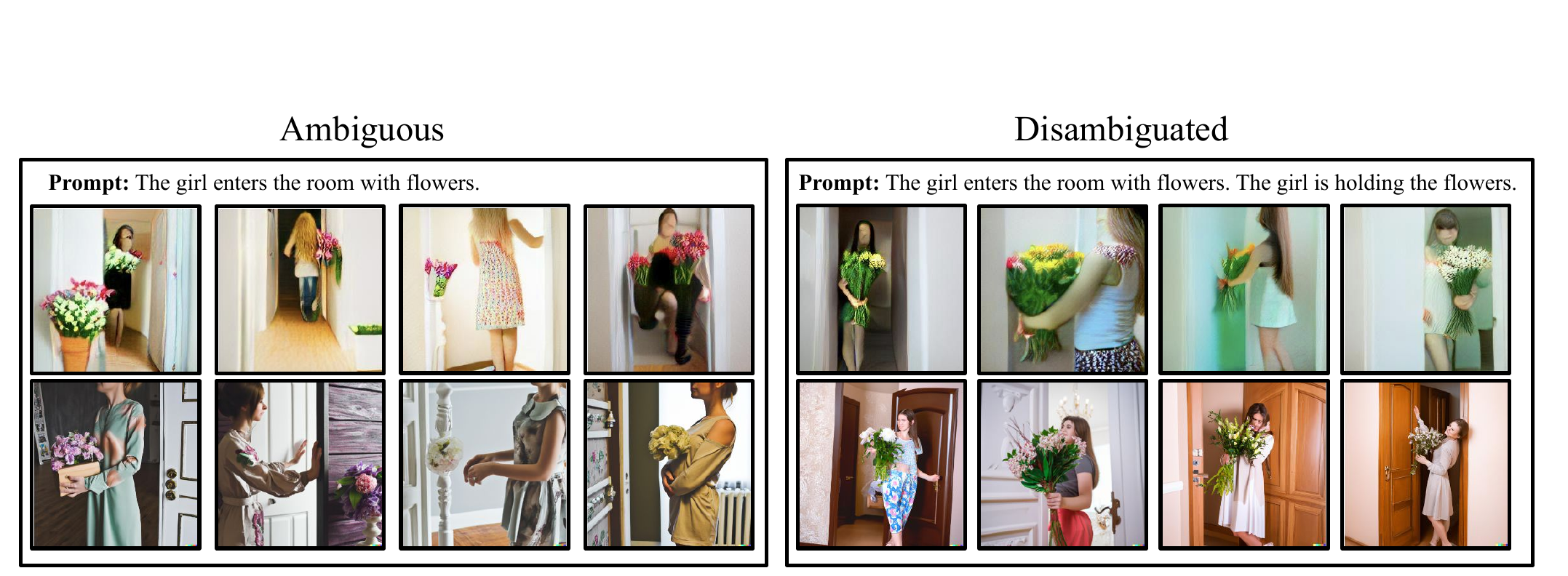}
    \vspace{-.07in}
    \caption{Qualitative examples generated by DALL-E Mega (top row) and OpenAI's DALL-E (bottom row).}
    \label{appendix:qualitative_overall}
\end{figure*}

We also demonstrate some qualitative results comparing the generated images qualitatively between ambiguous prompts provided to the system vs the disambiguated ones in Figure~\ref{appendix:qualitative_overall}.

\end{document}